\pdfoutput=1

\documentclass[11pt]{article}

\usepackage[final]{acl}

\usepackage{times}
\usepackage{latexsym}

\usepackage[T1]{fontenc}

\usepackage[utf8]{inputenc}

\usepackage{microtype}

\usepackage{inconsolata}
\usepackage{subfigure} 
\usepackage{amsmath}
\usepackage{stfloats}
\usepackage{multirow}
\usepackage{enumitem}
\usepackage{booktabs}
\usepackage{bbding}
\usepackage{amsfonts,amssymb}
\usepackage{diagbox}
\usepackage{newfloat}
\usepackage{listings}
\usepackage{times}  
\usepackage{helvet} 
\usepackage{courier}  
\usepackage{graphicx}
\usepackage{natbib}  
\usepackage{caption}
\usepackage{makecell}
\usepackage{xcolor}
\usepackage{pifont}
\usepackage{soul}
\usepackage{url}

\author{
 Meiqi~Chen$^{1,2}$\footnotemark[1], Yixin~Cao$^{3}$,
 Yan~Zhang$^{1,2}$$\footnotemark[2]$,  Chaochao~Lu$^{4}$$\footnotemark[2]$ \\ 
 $^1$State Key Laboratory of General Artificial Intelligence, Peking University, Beijing, China\\
 $^2$School of Intelligence Science and Technology, Peking University\\
 $^3$School of Computer Science, Fudan University\\
 $^4$Shanghai Artificial Intelligence Laboratory
 \\
\texttt{meiqichen@stu.pku.edu.cn}, 
\texttt{caoyixin2011@gmail.com}, \\ \texttt{zhyzhy001@pku.edu.cn},
\texttt{luchaochao@pjlab.org.cn}
}

\title{Quantifying and Mitigating Unimodal Biases in Multimodal Large Language Models: A Causal Perspective}

\begin{document}
\maketitle
\begin{abstract}

Recent advancements in Large Language Models (LLMs) have facilitated the development of Multimodal LLMs (MLLMs). Despite their impressive capabilities, MLLMs often suffer from over-reliance on unimodal biases (e.g., language bias and vision bias), leading to incorrect answers or hallucinations in complex multimodal tasks. To investigate this issue, we propose a causal framework to interpret the biases in Visual Question Answering (VQA) problems. Within this framework, we conduct an in-depth causal analysis to assess the causal effect of these biases on MLLM predictions.
Based on the analysis, we introduce 1) a novel \texttt{MORE} dataset with 12,000 challenging VQA instances requiring multi-hop reasoning and overcoming unimodal biases. 
2) a causality-enhanced agent framework  \texttt{CAVE} that guides models to comprehensively integrate information from different modalities and mitigate biases.
Our experiments show that MLLMs perform poorly on \texttt{MORE}, indicating strong unimodal biases and limited semantic understanding. However, when integrated with our \texttt{CAVE}, promising improvements in reasoning and bias mitigation can be seen.
These findings provide important insights for the development of more robust MLLMs and contribute to the broader goal of advancing multimodal AI systems capable of deeper understanding and reasoning.
Our project page is at {\small \url{https://github.com/OpenCausaLab/MORE}}.
\end{abstract}

\newcommand{\tabincell}[2]{\begin{tabular}{@{}#1@{}}#2\end{tabular}}
\renewcommand*{\thefootnote}{\fnsymbol{footnote}}
\footnotetext[1]{This work was done during her internship at Shanghai AI Laboratory.}
\footnotetext[2]{Corresponding author.}

\renewcommand*{\thefootnote}{\arabic{footnote}}
\setcounter{footnote}{0} 
\section{Introduction}
\label{sec:intro}

Following the success of Large Language Models (LLMs)~\cite{ouyang2022training, touvron2023llama}, Multimodal LLMs (MLLMs)~\cite{gpt2023openai, team2023gemini} have been proposed for various vision-language tasks~\cite{fu2023mme, liu2023mmbench}. Despite their promising results, it remains unclear if they truly understand images and text in the context of multimodal reasoning.

\begin{figure}
\centering  
\includegraphics[width=0.48\textwidth]{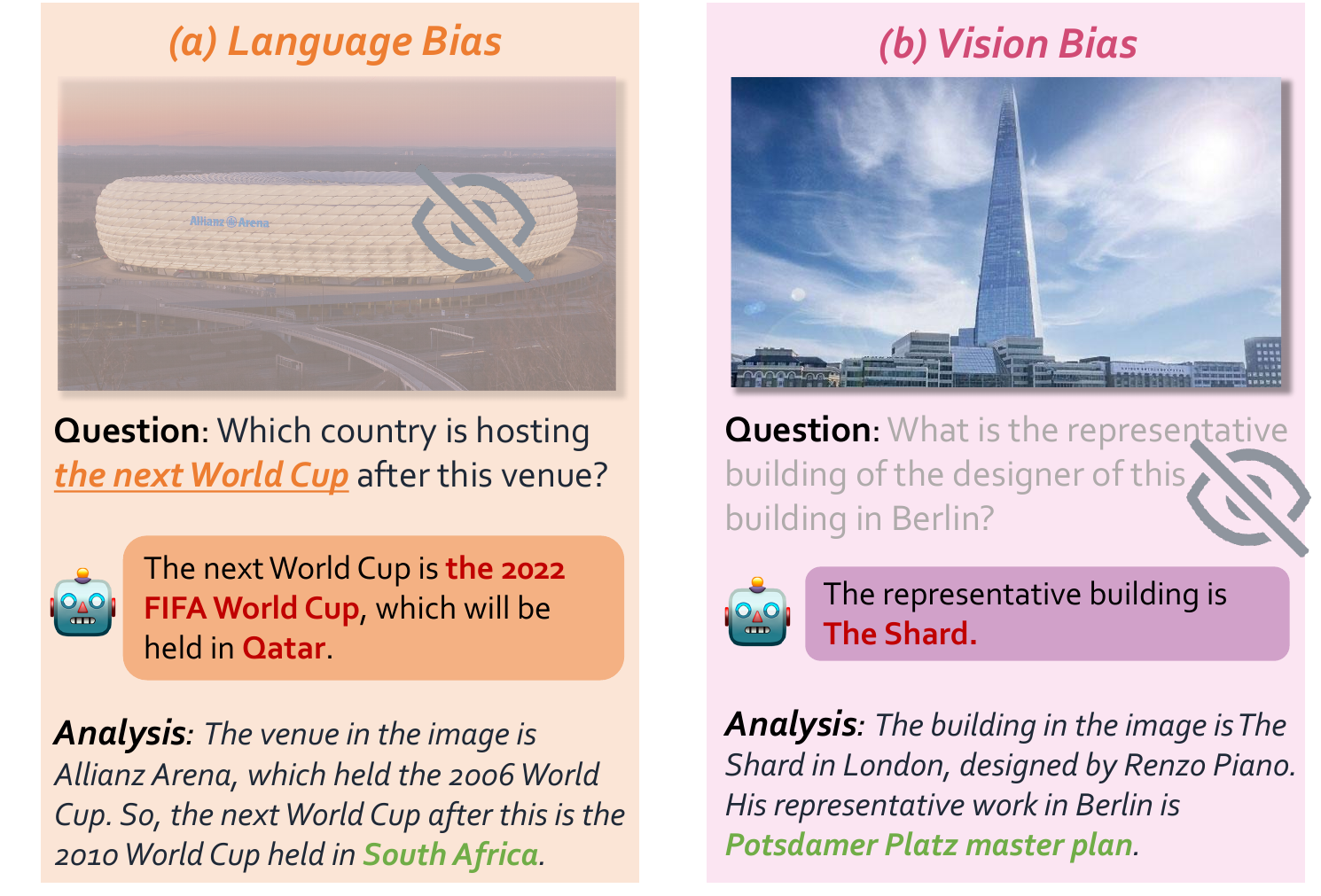}
\caption{Examples of over-reliance on unimodal biases. MLLMs (e.g., LLaVA) erroneously generate answers due to language bias~(indicated by the underlined text below the left image) and vision bias~(the right image).}
\label{fig:example}
\end{figure}

As shown in the knowledge-based Visual Question Answering (VQA) problems of Figure~\ref{fig:example}, when prompted with ``\emph{Which country is hosting the next World Cup after this venue?}'' MLLMs such as GPT-4V~\cite{gpt2023openai} and LLaVA~\cite{liu2023visual} may capture the language bias of ``\emph{the next World Cup}'' and think that the next World Cup will be ``\emph{the 2022 FIFA World Cup held in Qatar}'' (which is also outdated knowledge), while ignoring the exact venue presented in the image. Similarly, when presented with an image of ``\emph{The Shard}'' in London, MLLM directly identifies ``\emph{The representative building is The Shard}'' influenced by vision bias, overlooking the specific constraint ``\emph{in Berlin}'' mentioned in the question. These inherent issues pose significant challenges to the reasoning capabilities of MLLMs, particularly when faced with more complex questions. 

To investigate the issue of \textbf{M}LLMs' \textbf{O}ver-\textbf{RE}liance (\texttt{MORE}) on such unimodal biases, we propose a causal framework to interpret and quantify language and vision biases.
We begin by defining a causal graph of MLLM's prediction on VQA problems, built on key causal factors like images and questions. Then, we identify a set of interventions in the context of VQA problems, thereby ascertaining the causal effect of unimodal biases on MLLM predictions via $do$-calculus~\cite{pearl1995causal}. This allows us to evaluate the sensitivity and robustness of MLLMs against unimodal biases.

\begin{table*}
    \renewcommand
    \arraystretch{1.0}
    \centering
    \small
    \setlength{\tabcolsep}{4pt}
    \begin{tabular}{l|c|c|c|c|c|c}
    \toprule
        \bf{Datasets} & \textbf{\tabincell{c}{Knowledge-\\based}} &\bf{\tabincell{c}{Multi-hop \\ Reasoning}}     &{\bf{Answer Type}}  &\bf{\tabincell{c}{Unimodal Biases \\ Evaluation}}
        & \bf{Rationale}
        &\bf{\# Size}\\ 
 \midrule
         Visual7W ~\cite{zhu2016visual7w}  &\textcolor{red}{\ding{55}} &\textcolor{red}{\ding{55}} &Open-ended &\textcolor{red}{\ding{55}} &\textcolor{red}{\ding{55}} & 327.9K  \\
         VQA (v2)~\cite{goyal2017making}  &\textcolor{red}{\ding{55}} &\textcolor{red}{\ding{55}} &Open-ended &\textcolor{red}{\ding{55}} &\textcolor{red}{\ding{55}} & 1.1M  \\
        \midrule
         FVQA~\cite{wang2017fvqa}  &\textcolor{green}{\ding{51}} &\textcolor{red}{\ding{55}} &Open-ended&\textcolor{red}{\ding{55}} &\textcolor{green}{\ding{51}}  & 5.8K \\
         OKVQA~\cite{marino2019ok}  &\textcolor{green}{\ding{51}} &\textcolor{red}{\ding{55}} &Open-ended &\textcolor{red}{\ding{55}} &\textcolor{red}{\ding{55}}&14K   \\
         S3VQA~\cite{jain2021select}  &\textcolor{green}{\ding{51}} &\textcolor{red}{\ding{55}} &Open-ended&\textcolor{red}{\ding{55}} &\textcolor{red}{\ding{55}}&7.5K   \\
         A-OKVQA~\cite{schwenk2022okvqa}  &\textcolor{green}{\ding{51}} &\textcolor{red}{\ding{55}} &Multi-choice &\textcolor{red}{\ding{55}} &\textcolor{green}{\ding{51}}&23.7K   \\
         INFOSEEK~\cite{chen-etal-2023-pre-trained}  &\textcolor{green}{\ding{51}} &\textcolor{red}{\ding{55}} &Open-ended &\textcolor{red}{\ding{55}} &\textcolor{red}{\ding{55}} &1.4M  \\
         \midrule
         \bf{\texttt{MORE}} (Ours)  &\textcolor{green}{\ding{51}} &\textcolor{green}{\ding{51}} &Multi-choice &\textcolor{green}{\ding{51}} &\textcolor{green}{\ding{51}}&12K   \\
    \bottomrule
    \end{tabular}
        \caption{
        Comparison of \texttt{MORE} with other VQA datasets, highlighting its incorporation of external knowledge, multi-hop reasoning, unimodal bias evaluation, and rationale for interpretability. 
        }
    \label{tab:benchmark}
\end{table*}

Based on the above causal analysis, we curate a novel dataset termed \texttt{MORE}, comprising 12,000 VQA instances. This dataset advances existing VQA datasets by introducing a dedicated evaluation of unimodal biases.
We adopt a Multiple Choice Question (MCQ) format to facilitate the evaluation, where each instance consists of an image, a question, and four candidate options. The image is sourced from an existing VQA dataset~\cite{chen-etal-2023-pre-trained}. For question and option curation, we incorporate a knowledge graph (KG)~\cite{wang2021kepler}, allowing us to simulate MLLMs to navigate potential paths within the causal graph. Specifically, the options consist of one correct answer, and three distractors targeting language bias, vision bias, and multi-hop reasoning, respectively.
We also provide the reasoning path, designated as \textit{causal rationale}, in the KG for each instance, offering interpretability for evaluation.
As summarized in Table~\ref{tab:benchmark}, compared to existing VQA datasets, \texttt{MORE} features better comprehensiveness.

Furthermore, motivated by the causal analysis, we propose \texttt{CAVE}, a causality-enhanced method to mitigate unimodal biases in MLLMs. \texttt{CAVE} encompasses a diverse set of workflows, including question decomposition, causality-based enhanced self-reflection, external knowledge retrieval, and answer verification. This framework guides models in explicitly and comprehensively integrating information from multiple modalities while helping to prevent biased decision-making and the selection of incorrect shortcuts.

Through extensive experiments on \texttt{MORE} with several leading MLLMs, we observe that: 1) most MLLMs perform much poorly on \texttt{MORE}, showing strong reliance on unimodal biases and low robustness to disturbances; 2) MLLMs still struggle to achieve precise semantic understanding in multimodal reasoning; 3) With our propose method \texttt{CAVE}, we can mitigate the unimodal bias in MLLMs, yet it still falls short of the ideal. This indicates that addressing the unimodal bias in MLLMs is a highly challenging issue that merits further exploration.
Overall, our main contributions are as follows:
\begin{itemize}[leftmargin=*, itemsep=0pt, parsep=0pt]
\item We propose a causal framework to interpret and quantify the biases in VQA problems.
\item We construct a new dataset, \texttt{MORE}, which requires multi-hop reasoning and overcoming biases, demonstrating superior comprehensiveness to existing VQA datasets.
\item We conduct extensive experiments on \texttt{MORE} and propose a causality-enhanced method \texttt{CAVE} to mitigate the unimodal biases, providing insights for future work.
\end{itemize}

\section{A Causal Framework}
\label{sec:causal}
\begin{figure*}
\centering  
\includegraphics[width=1.0\textwidth]{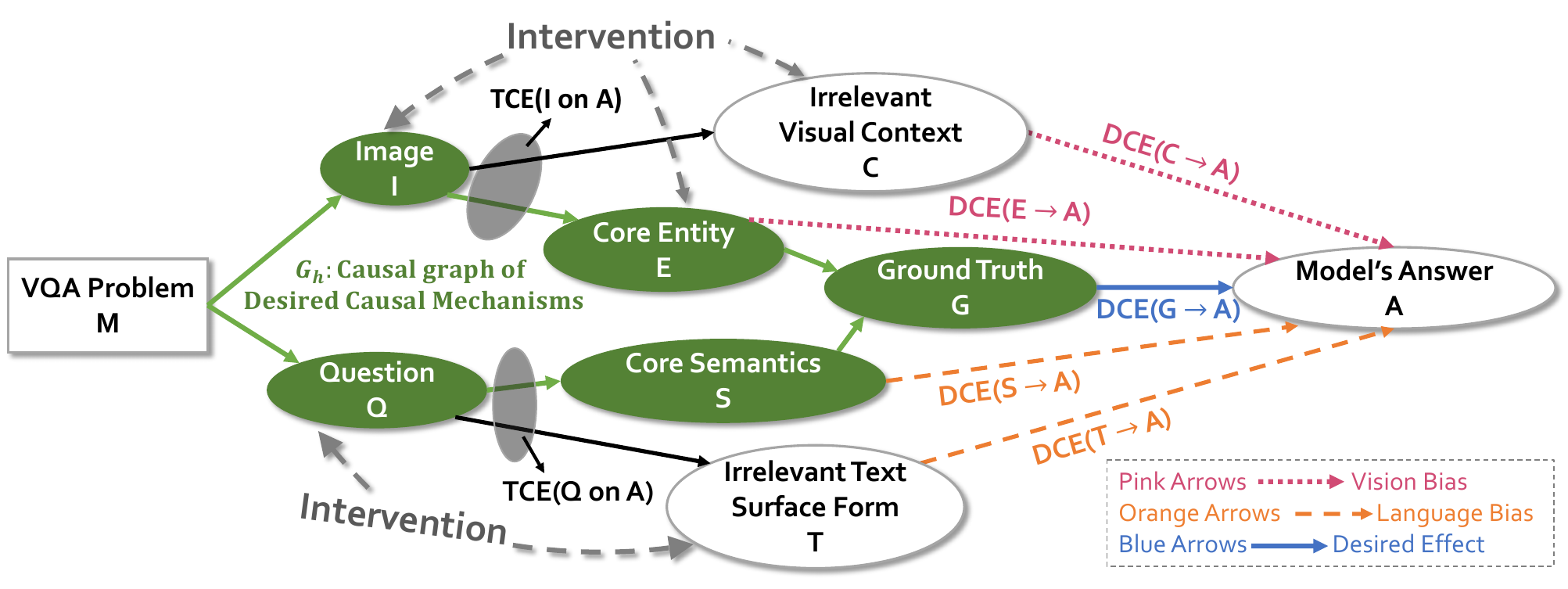}
\caption{Causal graph of MLLM's Prediction on VQA problems. We use the green subgraph $G_{h}$ to represent the desired causal mechanisms and compare it with the undesired effects of unimodal biases. 
We quantify the causal effects of each factor by performing controlled interventions of the images ($I, E, C$) and of the questions ($Q,T$).}
\label{fig:causal}
\end{figure*}
Inspired by \citet{stolfo-etal-2023-causal}, we introduce a causal graph for MLLM predictions on VQA problems, highlighting language and vision biases and assessing their causal effects via controlled interventions \cite{pearl1995causal}.
\subsection{Problem Setup}
We consider an entity-centric VQA problem, $M$, involving a question $Q := (S, T)$ and an image $I := (E, C)$. Here, $S$ is the core semantic content of $Q$, and $T$ is the textual form unrelated to $Q$'s main meaning. $I$ includes the primary entity $E$ and surrounding visual context $C$. The model outputs an answer, $A$. Notation: lowercase letters denote instances of corresponding uppercase variables.

\subsection{Causal Graph of MLLM Predictions}
\label{subsec:causal_graph}
Inspired by human cognitive processes~\cite{zellers2019recognition, stolfo-etal-2023-causal}, we formulate the causal mechanisms in human problem-solving for a VQA problem $m$: ${s}=f_{c_{1}}({q}), \ {e}=f_{c_{2}}({i}),\  g=f_{c_{3}}({s, e})$. This involves decoding the question $q$ to extract its semantic meaning $s$ through cognitive process $f_{c_{1}}$, and identifying the core entity $e$ from the image $i$ through $f_{c_{2}}$. These are combined by $f_{c_{3}}$ to produce the result $g$, as shown in the green subgraph $G_{h}$ of Figure~\ref{fig:causal}. In contrast, the model's approach to the same VQA problem employs $a=f_{b}({q}, {i})$, a black-box function where it is unclear how question and image inputs are utilized and interact to form the prediction $a$.
For deeper analysis, we illustrate possible causal mechanisms in Figure~\ref{fig:causal}. Notable mechanisms include:

\paragraph{Language Bias}  
The model may directly process the question $Q$ in two ways: by focusing on the core semantics $S$ via ${Q} \rightarrow S \rightarrow A$, or on the irrelevant part $T$ via ${Q} \rightarrow T \rightarrow A$. Both pathways lead to \emph{language bias}, e.g., the focus on ``the next World Cup '' in Figure~\ref{fig:example} (a).

\paragraph{Vision Bias} The model may attend directly to the entity $E$ of the image $I$ via $I\rightarrow {E} \rightarrow
A$, or to the irrelevant part $C$ via $I\rightarrow {C} 
\rightarrow A$. Both pathways
lead to the emergence of \emph{vision bias}, e.g., the focus on the ``\emph{The Shard}'' entity in Figure~\ref{fig:example} (b).

\paragraph{Desired Causal Mechanisms}  
Correct reasoning in VQA problems hinges on understanding causal mechanisms, as depicted in Figure~\ref{fig:causal}. The subgraph $G_{h}$ represents comprehension of how image and question jointly affect the ground-truth result ${G}$, through ${E} \rightarrow G$ and ${S} \rightarrow G$. This understanding should lead to model predictions that are sensitive and robust to changes in ${G}$, namely ${G} \rightarrow A$, with no spurious effects on $A$ unless it passes the mediator $G$. Model performance could therefore be evaluated by its: 
1) \emph{Sensitivity}, which assesses how well the model adapts to changes in the correct answer, i.e., $A$ responds to changes in $G$. 2) \emph{Robustness}, which measures resistance to unimodal biases,  e.g., $C \rightarrow A$ and $T \rightarrow A$, where minimal bias effects imply higher robustness to input changes that do not alter the ground-truth answer.

\subsection{Causal Analysis of VQA Biases}
\label{subsec:causal}
We adopt the controlled interventions as outlined by~\citet{pearl1995causal} to quantify the causal effects of questions and images on model predictions.
\paragraph{Causal Interventions for VQA}
\textbf{1)} \textit{Interventions on $Q$}. The question $Q$ can be modified in two ways:
(i) altering both $S$ and $T$ , or
(ii) altering $T$ but keeping $S$ unaffected. 
\textbf{2)} \textit{Interventions on $I$}. The image $I$ can be replaced with an alternative image $I^{\prime}$ in three ways:
(i) altering both $E$ and $C$ , or
(ii) altering $C$ but keeping $E$ unaffected, or
(iii) altering $E$ but keeping $C$ unaffected.
Note that we do not solely alter $S$ within $Q$, because it is not feasible to intervene on the core semantics $S$ of a question without affecting the surface text $T$ of it.

\paragraph{Formulation of Causal Effects}
We assess the causal effects using $\operatorname{do}\left(X: x \rightarrow x^{\prime}\right)$, where $X \in \{Q, T, I, C, E \}$ is a factor in the VQA problem $M$. The pre-intervention probability distribution $\mathbb{P}(A \mid I, Q)$ is denoted as $P$, and the post-intervention distribution as $P'$. Following  \citet{pearl1995causal}, the causal effect is evaluated using a distance metric $\delta$, as $\operatorname{CE}=\delta(P, P')$, where $ \operatorname{CE}$ denotes the causal effect. It can refer to 1) the total causal effect ($ \operatorname{TCE}$), signifying the joint effect across all causal paths from one variable to another; or 2) the direct causal effect ($ \operatorname{DCE}$), indicating the effect of the directed causal path devoid of intermediary variables~\cite{pearl2022direct}.
Following \citet{stolfo-etal-2023-causal}, we assess the causal effect of factor $X$ on the model's answer $A$ by comparing the change in predicted results, $\delta_{\mathrm{cp}}(P, P') := \mathbb{I}(a \neq a')$, where $a=\arg \max_x P(x)$ and $a'=\arg \max_{x} P'(x)$, with $\mathbb{I}$ indicating a change in the answer.

\paragraph{Causal Effects of Questions}
We assess $\operatorname{TCE}$ of a question $Q$ on an answer $A$ by intervening on $Q$:
\begin{equation}\label{eq:qtce}
\begin{aligned}
& \operatorname{TCE}({Q} \text { on } A):=\mathbb{E}_{{q}^{\prime} \sim \mathbb{P}({Q})}\left[\delta\left(P, P^{\prime}\right)\right], \\
& \text { where }P^{\prime}=\mathbb{P}\left(A \mid {I}, \operatorname{do}\left({Q}={q}^{\prime}\right)\right).
\end{aligned}
\vspace{-5pt} 
\end{equation}

This $\operatorname{TCE}$ contains two different types of paths that show how $Q$ affects $A$, as illustrated in Figure~\ref{fig:causal}:
\textbf{1)} The intended decision-making pathway: ${Q} \rightarrow S\rightarrow G \rightarrow A$, responding to changes in the ground truth. \textbf{2)} Potential spurious correlations, e.g., ${Q} \rightarrow T \rightarrow A$, where the model may depend on certain linguistic patterns from training data.

Maintaining the core semantics $S$ constant, we can assess $\operatorname{DCE}$ of the textual surface $T$ on $A$:
\begin{equation}\label{eq:qdce}
\begin{aligned}
& \operatorname{DCE}({T} \rightarrow A):=\mathbb{E}_{{q}^{\prime} \sim \mathbb{P}({Q} \mid S)}\left[\delta\left(P, P^{\prime}\right)\right], \\
& \text { where } P^{\prime}=\mathbb{P}\left(A \mid {I}, \operatorname{do}\left({Q}={q}^{\prime}\right)\right).
\end{aligned}
\vspace{-5pt} 
\end{equation}

As discussed in ``Causal Interventions for VQA'', solely intervening on $S$ without altering $T$ is impractical. However, Understanding $S$'s causal impact on $A$ is feasible by comparing two known quantities: $\operatorname{TCE}({Q} \text { on } A)$ and $\operatorname{DCE}({T} \rightarrow A)$.

\paragraph{Causal Effects of Images}
The causal structure of images mirrors that of questions, as shown in Figure~\ref{fig:causal}. Thus, we can derive $\operatorname{TCE}$ of image $I$ on the answer as follows:
\begin{equation}
\begin{aligned}
& \operatorname{TCE}({I} \text { on } A):=\mathbb{E}_{{i}^{\prime} \sim \mathbb{P}({I})}\left[\delta\left(P, P^{\prime}\right)\right], \\
& \text { where }P^{\prime}=\mathbb{P}\left(A \mid {Q}, \operatorname{do}\left({I}={i}^{\prime}\right)\right).
\end{aligned}
\vspace{-5pt} 
\end{equation}

Likewise, maintaining $E$ constant during each intervention on $I$ allows us to quantify the $\operatorname{DCE}$ of the irrelevant visual context $C$ on $A$:
\begin{equation}
\begin{aligned}
& \operatorname{DCE}({C} \rightarrow A):=\mathbb{E}_{{i}^{\prime} \sim \mathbb{P}({I} \mid E)}\left[\delta\left(P, P^{\prime}\right)\right], \\
& \text { where } P^{\prime}=\mathbb{P}\left(A \mid {Q}, \operatorname{do}\left({I}={i}^{\prime}\right)\right).
\end{aligned}
\vspace{-4pt} 
\end{equation}

Maintaining $C$ and $G$ constant during each intervention on $I$ instead allows us to quantify the $\operatorname{DCE}$ of the core entity $E$ on $A$:
\begin{equation}\label{eq:ieadce}
\begin{aligned}
& \operatorname{DCE}({E} \rightarrow A):=\mathbb{E}_{{i}^{\prime} \sim \mathbb{P}({I} \mid C, G)}\left[\delta\left(P, P^{\prime}\right)\right], \\
& \text { where } P^{\prime}=\mathbb{P}\left(A \mid {Q}, \operatorname{do}\left({I}={i}^{\prime}\right)\right).
\end{aligned}
\vspace{-4pt} 
\end{equation}

Overall, calculating $\operatorname{TCE}$ helps us assess models' sensitivity (response to changes in ground truth), while $\operatorname{DCE}$ evaluates its robustness (stability of predictions against spurious correlations).

\section{the \texttt{MORE} Dataset}
\label{subsec:MMQA}
\begin{figure*}[!htbp]
\centering  
\includegraphics[width=1.0\textwidth]{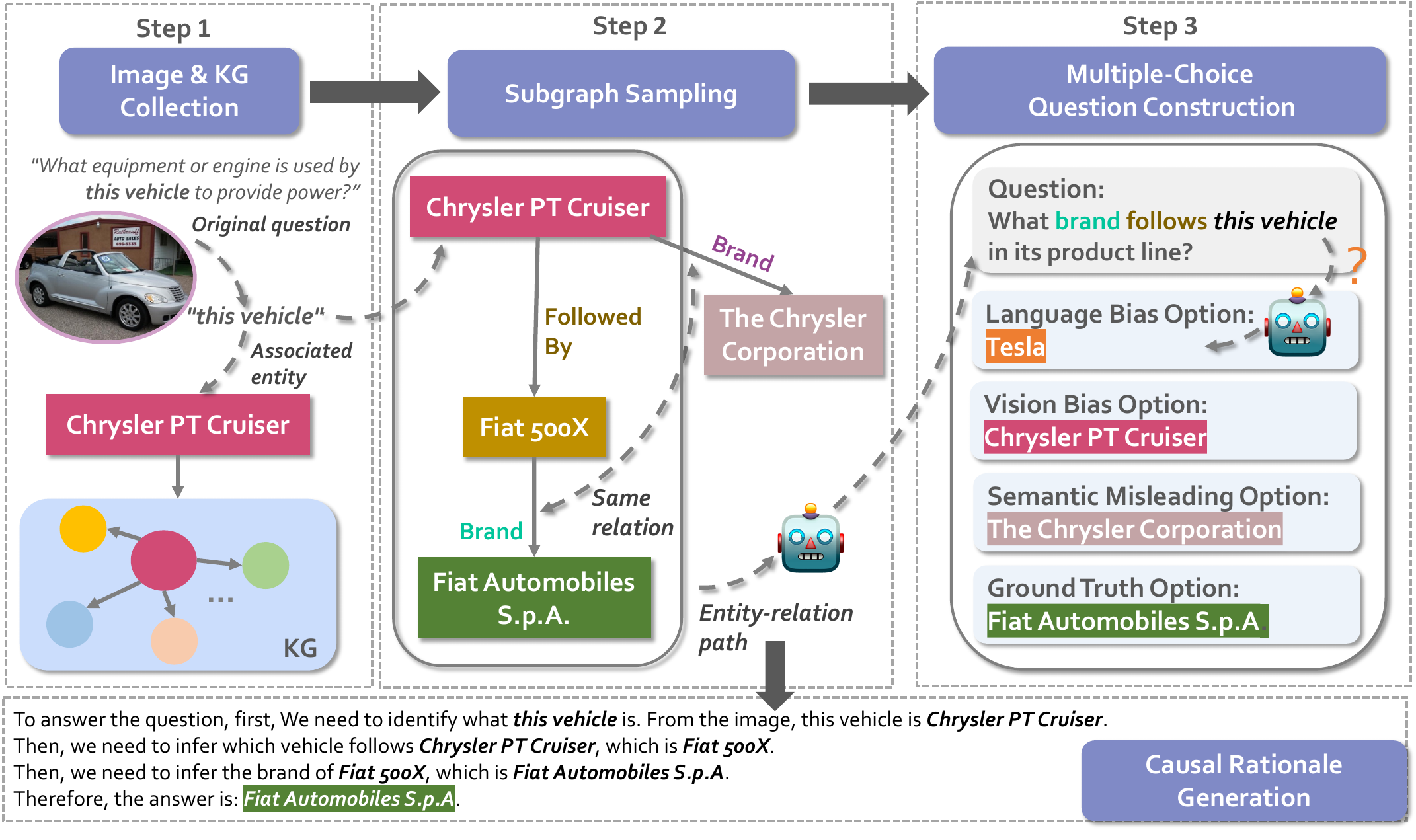}
\caption{Our framework for generating data of \texttt{MORE}. We first prepare the image source and link the visual entity in a knowledge graph. Then, motivated by the visual and language bias analysis through the causal lens, we construct multiple-choice questions that require MLLMs to overcome unimodal biases and conduct multi-hop reasoning in a sampled subgraph. We also generate the causal (reasoning) rationale for each instance to provide interpretability.}
\label{fig:question_generation}
\end{figure*}

From Section \ref{subsec:preparatory} to \ref{subsec:stat}, we construct the \texttt{MORE} 
 dataset that exploits the unimodal biases defined in Section \ref{subsec:causal_graph} and requires multi-hop reasoning.
In Section \ref{subsec:calculation}, we discuss how to quantify the causal effects of images and questions (defined in Section \ref{subsec:causal}) using our constructed dataset. 

\subsection{Preparatory Steps}
\label{subsec:preparatory}
\paragraph{Images and Knowledge Graph Collection}
We begin with an existing VQA dataset, INFOSEEK~\cite{chen-etal-2023-pre-trained}, which links image entities to Wikipedia information, requiring a VQA model to answer related questions. For instance in Figure~\ref{fig:question_generation}, the entity ``\emph{Chrysler PT Cruiser}'' prompts the question, ``\emph{What equipment or engine is used by this vehicle to provide power?}'' Here, terms like ``\emph{Chrysler PT Cruiser}'' and ``\emph{this vehicle}'' all refer to the same entity in the image.
We then identify all $n$-order neighbors ($n\in \{1,2\}$) of the associated entity within a knowledge graph (KG), Wikidata5M~\cite{wang2021kepler}, which is built upon Wikipedia data.

\paragraph{Subgraph Sampling}
Then, we identify a subgraph of an entity and its $n$-order neighbors in the KG. We filter paths in this subgraph that meet two criteria: 1) \emph{Uniqueness of Paths}: the path from the associated entity to the selected neighbor is unique, and 2) \emph{Shared-Type Relations}: they share a unique relation pointing to different entities. These criteria guarantee the uniqueness of the correct answer and introduce interference to challenge MLLMs' reasoning ability. For example, the filtered path in Figure~\ref{fig:question_generation} shows ``\emph{Fiat 500X}'' linked to ``\emph{Chrysler PT Cruiser}'' by a unique ``\emph{followed by}'' relation, and both connected to their respective companies via a ``\emph{brand}'' relation. This forms the multi-hop query: ``\emph{Chrysler PT Cruiser}'' $\stackrel{\text{followed by}}{\longrightarrow }$``\emph{Fiat 500X}'' $\stackrel{\text{brand}}{\longrightarrow }$ ``\emph{Fiat Automobiles S.p.A}''. Entities in the subgraph may act as distractors to challenge MLLMs' reasoning, which will be discussed in the next section.

\subsection{Multiple-Choice Question Construction}
\label{subsec:mc_question}
We detail the multiple-choice questions construction with four options, guided by the unimodal bias definition in Section \ref{subsec:causal_graph}.
\paragraph{Question Generation}
We generate questions by analyzing entity-relation paths within a specified subgraph, converting these paths into fluent, coherent queries with an LLM. We utilize the in-context learning (ICL) technique~\cite{brown2020language} and a standardized prompt (shown in Appendix~\ref{app:prompt_question}) to ensure quality. Among various LLMs tested, ChatGPT was selected for its superior multi-hop question generation. To avoid information leakage, entity names in the question are anonymized as ``this <ENTITY\_NAME>''.
For instance, a final generated question asks ``\emph{What brand follows this vehicle in its product line?}'' in Figure~\ref{fig:question_generation}.

\paragraph{Language Bias Option}
Language bias occurs when models overly attend to question-related information via pathways like ${Q} \rightarrow T \rightarrow A$. To assess this, we simulate conditions where only the question is provided and evaluate MLLMs' responses. We use the answers of GPT-4 (i.e., the \emph{text-only} version of GPT-4V), to ensure uniform final options across MLLMs. For example, in Figure~\ref{fig:question_generation}, GPT-4 reponds ``\emph{Tesla}'' to the question. 
Additional results using different models for generating language bias options are detailed in Section~\ref{subsec:exp_causal} and the prompt template is available in Appendix~\ref{app:prompt_language}.

\paragraph{Vision Bias Option}
Vision bias occurs when visual information dominates (e.g., via $I\rightarrow {E} \rightarrow
A$). We use the visually associated entity name (e.g., ``\emph{Chrysler PT Cruiser}'') as an option, to see if the model directly selects it upon encountering an option that aligns with the visual information.

\paragraph{Semantic Misleading Option}
We introduce a semantic misleading option, such as ``\emph{The Chrysler Corporation}'', to challenge MLLMs' multi-hop reasoning. This option refers to the entity that is pointed by the relation commonly owned by both the associated entity and its sampled neighbor.
For example in Figure~\ref{fig:question_generation}, upon encountering a question containing ``\emph{brand}'' and ``\emph{Chrysler PT Cruiser}'', MLLMs might simply output a direct answer (e.g., ``\emph{The Chrysler Corporation}''), ignoring other constraints in the question (e.g., ``\emph{followed by}''), hence struggling to choose the correct answer (e.g., ``\emph{Fiat Automobiles S.p.A.}'').

\paragraph{Ground Truth Option} 
Corresponding to the causal path
via ${E} \rightarrow G$ and ${S} \rightarrow G$, this option represents the final entity in the entity-relation path (e.g., ``\emph{Fiat Automobiles S.p.A.}'').
Finally, we check and ensure that each option is unique to prevent overlap samples.

\paragraph{Causal Rationale Generation}
Using entity-relation paths, we generate a causal rationale that aids in answering questions through a heuristic rule-based approach. As shown in Figure~\ref{fig:question_generation}, this process begins with the associated entity and proceeds step-by-step to the ground truth. These rationales help confirm the accuracy of MLLMs' reasoning and improve their interpretability. They also can contribute to fine-tuning MLLMs for better multi-hop reasoning.

\subsection{Dataset Statistics and Quality Analysis}
\label{subsec:stat}
\paragraph{Statistics of Different Hops}
We automatically generate training data from INFOSEEK's train set, and development/test data from its validation set, as shown in Table \ref{tab:dataset_stat} in Appendix.
Focusing on questions with 2-hop and 3-hop depths, we set $n=1, 2$ respectively. We avoid longer-hop questions to prevent potential ambiguity and complexity.

\paragraph{Question Distribution}
We categorize the generated questions into distinct types based on their starting n-grams in Figure~\ref{fig:question_distribution} in the Appendix. The \texttt{MORE} dataset showcases an extensive lexical diversity in the questions generated. 

\paragraph{Question Quality}
We analyze the lexical diversity and fluency of the generated questions, with baselines and metrics detailed in Appendix~\ref{app:question_quality}. From Figure~\ref{fig:question_quality}, \texttt{MORE} shows superiority in lexical diversity and fluency, even compared to human-generated datasets.

\paragraph{Human Evaluation}
Our human evaluation confirms the high quality of generated questions and rationales, with 91\% of questions and 98\% of rationales deemed valid by annotators (details are in Appendix~\ref{app:human_evaluation}).

\begin{figure}
\centering  
\includegraphics[width=0.48\textwidth]{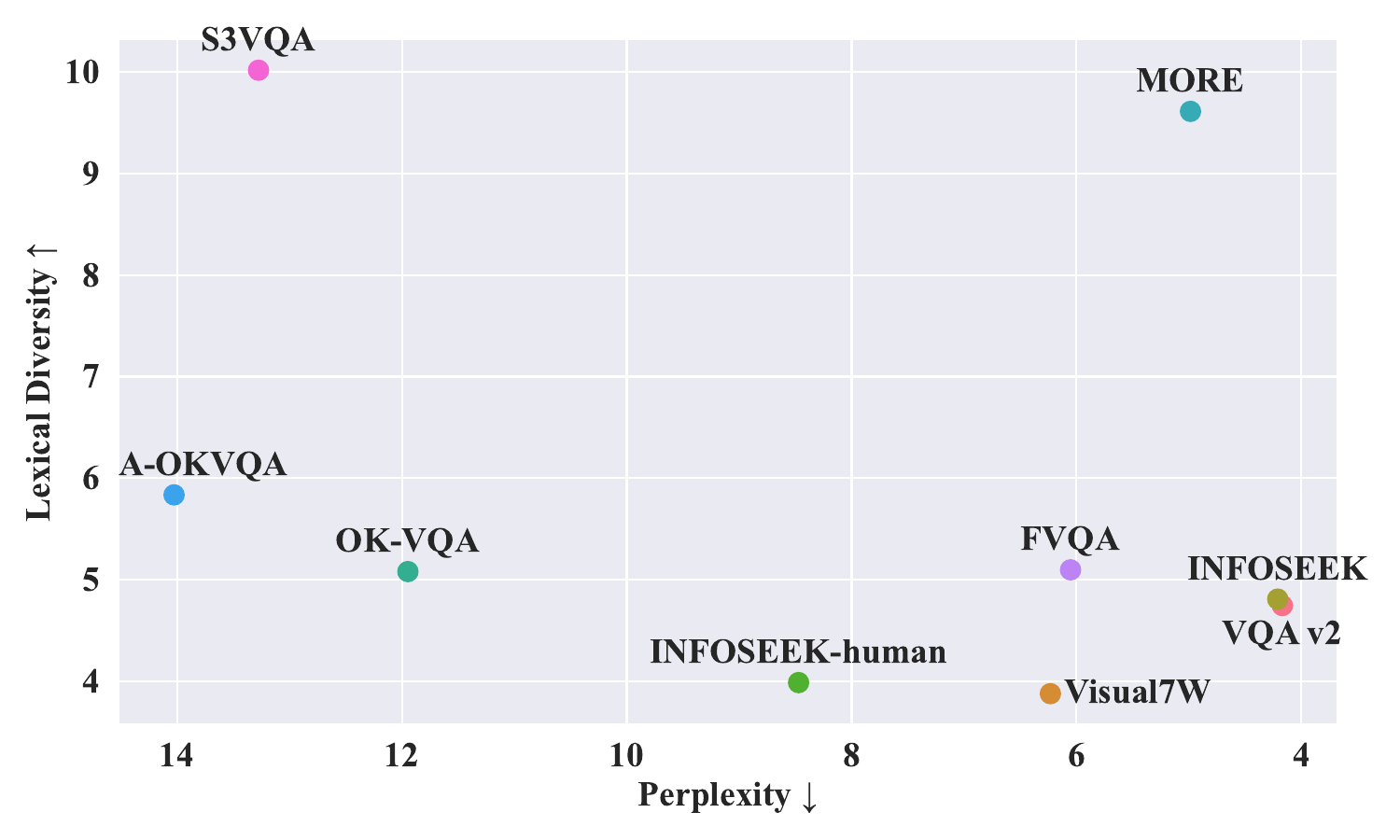}
\caption{Question quality of \texttt{MORE} compared to other VQA datasets in terms of lexical diversity and fluency.}
\label{fig:question_quality}
\end{figure}

\subsection{Causal Effect Calculation on \texttt{MORE}}
\label{subsec:calculation}
Based on the analysis in Section \ref{subsec:causal}, we discuss how to quantify the causal effects of unimodal biases on model predictions using \texttt{MORE}.

\paragraph{Causal Effect of Questions}
When intervening on questions, we keep the images constant:
(1) For $\operatorname{TCE}({Q} \text { on } A)$, we alter the question to another from the dataset pertaining to the same entity, changing both its text and core semantics, which shifts the ground truth.
(2) For $\operatorname{DCE}({T} \rightarrow A)$, we let ChatGPT rephrase the question, altering its textual form but preserving its semantic meaning, resulting in no ground truth change post-intervention.

\paragraph{Causal Effect of Images}
When intervening on images, we keep the questions constant:
(1) For $\operatorname{TCE}({I} \text { on } A)$, we replace the image with another from the dataset corresponding to the same question but featuring different entities, altering both core entity and visual context to change the ground truth post-intervention.
(2) For $\operatorname{DCE}({C} \rightarrow A)$,
we replace the image with another that depicts the same entity, to keep the ground truth consistent after the intervention.  Besides, it is practically challenging to calculate $\operatorname{DCE}({E} \rightarrow A)$.
Because intervening on core entity $E$ while maintaining the constancy of the question $Q$, visual context $C$, and ground truth answer $G$ is hard to achieve, hence not considered in further analysis.

Given the impracticality of testing all perturbations of $T$ and $C$, we randomly select 100 samples for each type of intervention and compute the average effects to determine $\operatorname{TCE}$ and $\operatorname{DCE}$ in Section \ref{subsec:exp_causal}. Overall, a higher $\operatorname{TCE}$ indicates better sensitivity, while a lower $\operatorname{DCE}$ indicates better robustness. 

\section{\texttt{CAVE} for Bias Mitigation}
To mitigate unimodal biases in MLLMs and improve their reasoning capabilities, we propose \texttt{CAVE}, a causality-enhanced agent framework in this section.

For a given instance, \texttt{CAVE} starts by using a question decomposer to break down complex questions into simpler, step-by-step subquestions, explicitly avoiding spurious paths that may lead to incorrect answers. For each subquestion, \texttt{CAVE} uses a causality-enhanced reasoner to evaluate the correctness of the decomposed subquestion. Specifically, based on the aforementioned causal analysis in Section~\ref{sec:causal}, it assesses whether the decomposed subquestion changes under two conditions: 1) rephrasing the question text, or 2) replacing the image with another captured by the same entity from a different angle or at a different time. If it identifies that the decomposed subquestion needs to be altered, this indicates that the previous understanding is influenced by irrelevant factors or biases, failing to capture the true semantics. In such cases, it will enforce a new round of question decomposition until it determines that the current subquestions are appropriate and accurate.
Next, \texttt{CAVE} uses a verifier to strategically employ external tools and evaluate their outputs, acquiring the necessary context or information, such as image and text retrieval, to provide a precise answer. This iterative process of answering and verifying continues until all subquestions are resolved. By incorporating external knowledge, the final verified output integrates information from the entire reasoning process to provide a correct answer. A detailed illustration of \texttt{CAVE} and the prompt template is provided in Appendix~\ref{app:cave}.
\section{Experiments}
\begin{table*}
    \renewcommand
    \arraystretch{1.0}
    \centering
    \setlength{\tabcolsep}{1pt}
    \begin{tabular}{l|c|c|c|c|c}
    \toprule
         \bf{Model}&\bf{LLM}   &\bf{\# Params}& \textbf{Two-Hop, acc (\%)} & \textbf{Three-Hop, acc (\%)} & \textbf{Overall, acc (\%)} \\ \midrule
          Random &/&/  &25.0  &{25.0}  &{25.0} \\ \midrule
         InstructBlip &Vicuna &13B  &17.0  &16.2 &16.6\\
         mPLUG-Owl &Llama &7B  &12.4  &11.4 &11.9 \\
         LLaVA &Llama &13B  &20.8  &13.6  &17.5 \\
         Qwen-VL &Qwen &7B  &17.4  &15.6  &16.5 \\
         \midrule
         GPT-4v &-&- &17.3  &16.0  &16.5  \\
         GPT-4o &-&-  &18.0 &17.0  &17.5  \\
         GPT-4o + \texttt{CAVE} &-&-  &33.7 &27.3  &30.5 \\\midrule
         Gemini Pro (V)&-&-  &{25.4}  &20.7  &{22.3} \\
        Gemini Pro (V)  + \texttt{CAVE} &-&-  &\textbf{35.6}  &\textbf{28.8}  &\textbf{33.2} \\

    \bottomrule
    \end{tabular}
        \caption{MLLMs' results on the test set of \texttt{MORE}. We report the VQA accuracy (\%) under the multi-choice settings on two-hop, three-hop, and all data, respectively.``-'' denotes not released information.}
    \label{tab:main}
\end{table*}
\subsection{Experimental Setup}
\paragraph{Datasets}
We evaluate all test data from the \texttt{MORE} dataset using the \textbf{Multi-choice} settings, where MLLMs select answers from four provided options, with a random baseline accuracy of 25\%.

\paragraph{Baselines}
We evaluate various leading MLLMs on our \texttt{MORE} dataset in a zero-shot fashion, including three limited-access MLLMs: GPT-4v, GPT-4o~\cite{gpt2023openai}, and Gemini Pro Vision~\cite{team2023gemini},
and four open-source MLLMs: InstructBLIP (13B) ~\cite{dai2023instructblip}, mPLUG-Owl (7B)~\cite{ye2023mplug}, LLaVA (v1.5, 13B)~\cite{liu2023visual}, and Qwen-VL (7B)~\cite{bai2023qwen}
the details are in Appendix~\ref{app:baselines}. For consistent evaluation, we use standard accuracy metrics for all the models. We further quantify the causal effects of images and questions on model predictions following Section~\ref{subsec:exp_causal}.

\subsection{Evaluation Results}
The results of MLLMs on \texttt{MORE} are shown in Table~\ref{tab:main} and further exemplified in Appendix~\ref{app:case}. We observe that:

1) All baselines perform poorly on \texttt{MORE} (e.g., even the best-performing model, Gemini Pro Vision, achieves only 22.3\% accuracy, which does not surpass the random baseline), indicating MLLMs' vulnerability to biases.

2) Most models perform better on two-hop data than on three-hop data, suggesting that MLLMs' reasoning capabilities are challenged when the problems become more complex.

3) GPT-4v falls short versus Gemini Pro Vision, possibly because we use homologous GPT-generated distractors when constructing the language bias options in Section~\ref{subsec:mc_question}, which particularly challenges GPT-4v's judgment. This point is further analyzed in Section~\ref{subsec:exp_causal}.

4) Our proposed \texttt{CAVE} significantly enhances GPT-4o and Gemini Pro Vision on \texttt{MORE}, validating the effectiveness of our method. However, the relatively low absolute values indicate ongoing challenges related to biases, suggesting the need for further research efforts.

\subsection{Causal Analysis of VQA Biases}
\label{subsec:exp_causal}
\begin{figure}
\centering  
\includegraphics[width=0.48\textwidth]{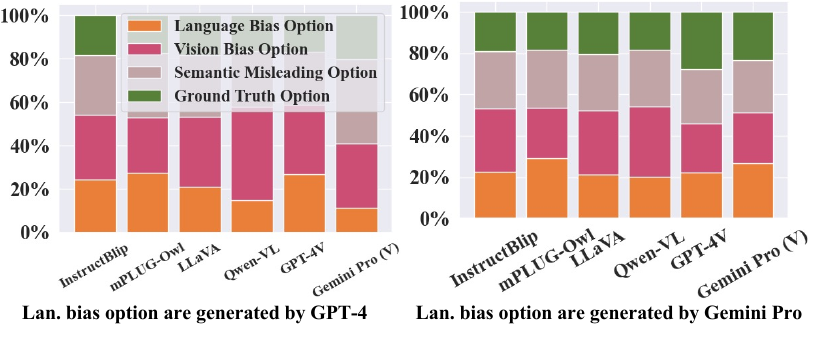}
\caption{Option distribution of MLLMs.}
\label{fig:option_distribution}
\end{figure}

In this subsection, we select several representative MLLMs and analyze their performance through a causal lens. 
\paragraph{Option Distribution}
In Figure~\ref{fig:option_distribution}, we show the option distribution of selected MLLMs. Here we mainly utilize language bias options generated by GPT-4v and Gemini Pro Vision for comparison because they explicitly provide corresponding text-only versions.
The observations are as follows:

1) \emph{Severe unimodal biases}. More than 40\% of the options show either language or vision bias in all models, underscoring the prevalence of unimodal biases.

2) \emph{Limited understanding ability}. Models' selection of semantically misleading options indicates some ability to combine visual and textual information, though not fully grasping the problem. This highlights the challenge our \texttt{MORE} dataset poses to current MLLMs. 

3) \emph{Obvious selection tendency}.  GPT-4v often incorrectly chooses language bias options generated from GPT-4 (i.e., the text-only version of GPT-4v). Switching to Gemini Pro (i.e., the text-only version of Gemini Pro Vision) shifts this trend, with GPT-4v's language bias selections decreasing and Gemini Pro's increasing.
These observations align with our prior analysis of GPT-4v. 

Please note that discrepancies may exist between the proportions of ground truth options presented here and the accuracy values reported in Table \ref{tab:main}, as some models' outputs may not align with the provided option format (e.g., mPLUG-Owl), thus affecting the count of valid answers.
\begin{figure}
\centering  
\includegraphics[width=0.48\textwidth]{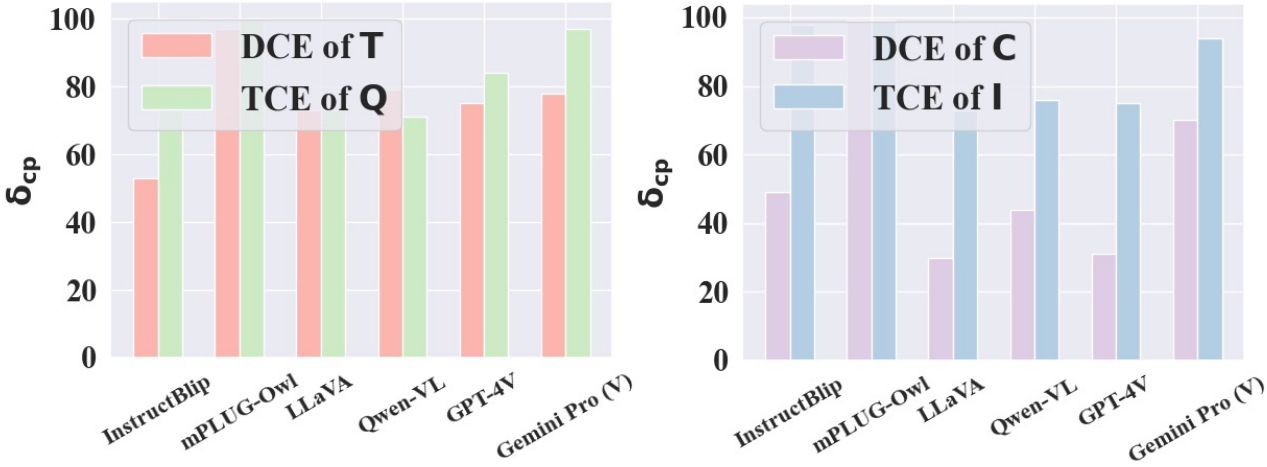}
\caption{Comparison of direct and total effects of image and question on prediction for MLLMs.} 
\label{fig:language_bias}
\end{figure}
\paragraph{Causal Effects of Images and Questions}
To further analyze the impact of unimodal biases on the model predictions, we assess the causal effects based on the discussion in Section~\ref{subsec:calculation}. As shown in Figure~\ref{fig:language_bias}, we select several representative MLLMs: 

1) Current MLLMs exhibit high sensitivity (high $\operatorname{TCE}$), a possible reason is that instruction tuning makes models sensitive to variations in input~\cite{stolfo-etal-2023-causal}.

2) However, the model's robustness is relatively low, as indicated by a high $\operatorname{DCE}$, meaning its predictions fluctuate even when the ground truth remains constant. This phenomenon becomes particularly pronounced when irrelevant text surface forms are introduced. This suggests that, although MLLMs can adapt well to answers, they exhibit a lack of robustness when exposed to interference, relying on spurious correlations rather than genuine causal features.
\section{Related Work}
\label{sec:related_work}

\paragraph{Multimodal Large Language Models (MLLMs)}
Recent advances in large language models (LLMs) have led to the emergence of MLLMs, which demonstrate exceptional performance in multimodal tasks~\cite{gpt2023openai, team2023gemini, liu2023visual, viscpm, rlhfv}. However, in contrast to the extensive evaluation of reasoning capabilities in LLMs~\cite{wei2022chain, chen-etal-2024-improving-large}, currently, the evaluation of MLLMs primarily emphasizes basic visual tasks \cite{liu2023mmbench, fu2023mme, lu2024gpt, chen2024cello}, with limited investigation into their reasoning capabilities.

\paragraph{Knowledge-based VQA Datasets}
Existing VQA datasets~\cite{wang2017fvqa, marino2019ok, chen-etal-2023-pre-trained} are limited by their focus on image-related information, lack of multi-hop reasoning, open-ended answers, and reasoning rationales. They also fail to measure the effect of language and vision biases. Our MORE dataset, outlined in Table \ref{tab:benchmark}, addresses these shortcomings by providing a more comprehensive assessment.

\paragraph{Language and Vision Biases in VQA}
Research reveals that some VQA models primarily depend on statistical priors from training data instead of genuinely comprehending image content~\cite{agrawal2018don}. These models exhibit language and vision biases; the former arises from strong correlations between specific questions and answers~\cite{abbasnejad2020counterfactual, zhu2020overcoming}, and the latter from frequent co-occurrences of textual and visual elements in the dataset~\cite{si2022language, gupta2022swapmix}. Recent efforts to address these biases mostly involve data augmentation~\cite{niu2021counterfactual}. Besides, several studies \cite{{rohrbach-etal-2018-object, parcalabescu-etal-2022-valse,  parcalabescu-frank-2023-mm}} use examples derived from language biases to assess object hallucination in models. However, these constructed examples are typically counterfactual, with the answer types limited to yes/no, essentially judging whether the constructed statements are correct or incorrect.

\section{Conclusion}
This paper presents a comprehensive approach to quantifying and mitigating the unimodal biases in MLLMs. Through our causal inference framework, we provide an in-depth analysis to assess the causal effects of such biases on the model's prediction in VQA problems. The introduced \texttt{MORE} dataset challenges MLLMs to engage in multi-hop reasoning and to overcome language and vision biases, thereby pushing the boundaries of their reasoning capabilities. Our proposed \texttt{CAVE} method demonstrates significant potential in enhancing the reasoning abilities of MLLMs.
\section*{Limitations}
Our current generation of rationales is based on heuristic rules. Previous works have demonstrated the effectiveness of incorporating rationales into instructions~\cite{wei2022chain}. Therefore, we believe that refining and polishing these rationales with an LLM (e.g., ChatGPT) could be beneficial.  Besides, the Wikidata5M dataset we employed was released in 2021, and some information in the knowledge graph may be outdated. Although we have made efforts to manually verify the test set and try to ensure it does not contain incorrect information, it is still inevitable that errors may occur within the extensive training data.

\section*{Acknowledgments}
We thank all the anonymous reviewers for their valuable feedback throughout the review process.
This work is supported in part by Ucap Cloud and the State Key Laboratory of General Artificial Intelligence.

\bibliography{custom}

\begin{thebibliography}{53}
\expandafter\ifx\csname natexlab\endcsname\relax\def\natexlab#1{#1}\fi

\bibitem[{Abbasnejad et~al.(2020)Abbasnejad, Teney, Parvaneh, Shi, and van~den Hengel}]{abbasnejad2020counterfactual}
Ehsan Abbasnejad, Damien Teney, Amin Parvaneh, Javen Shi, and Anton van~den Hengel. 2020.
\newblock \href {https://doi.org/10.1109/CVPR42600.2020.01006} {Counterfactual vision and language learning}.
\newblock In \emph{2020 {IEEE/CVF} Conference on Computer Vision and Pattern Recognition, {CVPR} 2020, Seattle, WA, USA, June 13-19, 2020}, pages 10041--10051. {IEEE}.

\bibitem[{Agrawal et~al.(2018)Agrawal, Batra, Parikh, and Kembhavi}]{agrawal2018don}
Aishwarya Agrawal, Dhruv Batra, Devi Parikh, and Aniruddha Kembhavi. 2018.
\newblock \href {https://doi.org/10.1109/CVPR.2018.00522} {Don't just assume; look and answer: Overcoming priors for visual question answering}.
\newblock In \emph{2018 {IEEE} Conference on Computer Vision and Pattern Recognition, {CVPR} 2018, Salt Lake City, UT, USA, June 18-22, 2018}, pages 4971--4980. {IEEE} Computer Society.

\bibitem[{Bai et~al.(2023{\natexlab{a}})Bai, Bai, Chu, Cui, Dang, Deng, Fan, Ge, Han, Huang et~al.}]{bai2023qwent}
Jinze Bai, Shuai Bai, Yunfei Chu, Zeyu Cui, Kai Dang, Xiaodong Deng, Yang Fan, Wenbin Ge, Yu~Han, Fei Huang, et~al. 2023{\natexlab{a}}.
\newblock Qwen technical report.
\newblock \emph{arXiv preprint arXiv:2309.16609}.

\bibitem[{Bai et~al.(2023{\natexlab{b}})Bai, Bai, Yang, Wang, Tan, Wang, Lin, Zhou, and Zhou}]{bai2023qwen}
Jinze Bai, Shuai Bai, Shusheng Yang, Shijie Wang, Sinan Tan, Peng Wang, Junyang Lin, Chang Zhou, and Jingren Zhou. 2023{\natexlab{b}}.
\newblock Qwen-vl: A frontier large vision-language model with versatile abilities.
\newblock \emph{arXiv preprint arXiv:2308.12966}.

\bibitem[{Brown et~al.(2020)Brown, Mann, Ryder, Subbiah, Kaplan, Dhariwal, Neelakantan, Shyam, Sastry, Askell, Agarwal, Herbert{-}Voss, Krueger, Henighan, Child, Ramesh, Ziegler, Wu, Winter, Hesse, Chen, Sigler, Litwin, Gray, Chess, Clark, Berner, McCandlish, Radford, Sutskever, and Amodei}]{brown2020language}
Tom~B. Brown, Benjamin Mann, Nick Ryder, Melanie Subbiah, Jared Kaplan, Prafulla Dhariwal, Arvind Neelakantan, Pranav Shyam, Girish Sastry, Amanda Askell, Sandhini Agarwal, Ariel Herbert{-}Voss, Gretchen Krueger, Tom Henighan, Rewon Child, Aditya Ramesh, Daniel~M. Ziegler, Jeffrey Wu, Clemens Winter, Christopher Hesse, Mark Chen, Eric Sigler, Mateusz Litwin, Scott Gray, Benjamin Chess, Jack Clark, Christopher Berner, Sam McCandlish, Alec Radford, Ilya Sutskever, and Dario Amodei. 2020.
\newblock \href {https://proceedings.neurips.cc/paper/2020/hash/1457c0d6bfcb4967418bfb8ac142f64a-Abstract.html} {Language models are few-shot learners}.
\newblock In \emph{Advances in Neural Information Processing Systems 33: Annual Conference on Neural Information Processing Systems 2020, NeurIPS 2020, December 6-12, 2020, virtual}.

\bibitem[{Cahyawijaya et~al.(2021)Cahyawijaya, Winata, Wilie, Vincentio, Li, Kuncoro, Ruder, Lim, Bahar, Khodra, Purwarianti, and Fung}]{cahyawijaya-etal-2021-indonlg}
Samuel Cahyawijaya, Genta~Indra Winata, Bryan Wilie, Karissa Vincentio, Xiaohong Li, Adhiguna Kuncoro, Sebastian Ruder, Zhi~Yuan Lim, Syafri Bahar, Masayu Khodra, Ayu Purwarianti, and Pascale Fung. 2021.
\newblock \href {https://doi.org/10.18653/v1/2021.emnlp-main.699} {{I}ndo{NLG}: Benchmark and resources for evaluating {I}ndonesian natural language generation}.
\newblock In \emph{Proceedings of the 2021 Conference on Empirical Methods in Natural Language Processing}, pages 8875--8898, Online and Punta Cana, Dominican Republic. Association for Computational Linguistics.

\bibitem[{Chen et~al.(2024{\natexlab{a}})Chen, Ma, Song, Cao, Zhang, and Li}]{chen-etal-2024-improving-large}
Meiqi Chen, Yubo Ma, Kaitao Song, Yixin Cao, Yan Zhang, and Dongsheng Li. 2024{\natexlab{a}}.
\newblock \href {https://doi.org/10.18653/v1/2024.acl-long.512} {Improving large language models in event relation logical prediction}.
\newblock In \emph{Proceedings of the 62nd Annual Meeting of the Association for Computational Linguistics (Volume 1: Long Papers)}, pages 9451--9478, Bangkok, Thailand. Association for Computational Linguistics.

\bibitem[{Chen et~al.(2024{\natexlab{b}})Chen, Peng, Zhang, and Lu}]{chen2024cello}
Meiqi Chen, Bo~Peng, Yan Zhang, and Chaochao Lu. 2024{\natexlab{b}}.
\newblock Cello: Causal evaluation of large vision-language models.
\newblock \emph{arXiv preprint arXiv:2406.19131}.

\bibitem[{Chen et~al.(2022)Chen, Hu, Chen, Verga, and Cohen}]{chen2022murag}
Wenhu Chen, Hexiang Hu, Xi~Chen, Pat Verga, and William Cohen. 2022.
\newblock \href {https://aclanthology.org/2022.emnlp-main.375} {{M}u{RAG}: Multimodal retrieval-augmented generator for open question answering over images and text}.
\newblock In \emph{Proceedings of the 2022 Conference on Empirical Methods in Natural Language Processing}, pages 5558--5570, Abu Dhabi, United Arab Emirates. Association for Computational Linguistics.

\bibitem[{Chen et~al.(2023)Chen, Hu, Luan, Sun, Changpinyo, Ritter, and Chang}]{chen-etal-2023-pre-trained}
Yang Chen, Hexiang Hu, Yi~Luan, Haitian Sun, Soravit Changpinyo, Alan Ritter, and Ming-Wei Chang. 2023.
\newblock \href {https://doi.org/10.18653/v1/2023.emnlp-main.925} {Can pre-trained vision and language models answer visual information-seeking questions?}
\newblock In \emph{Proceedings of the 2023 Conference on Empirical Methods in Natural Language Processing}, pages 14948--14968, Singapore. Association for Computational Linguistics.

\bibitem[{Chiang et~al.(2023)Chiang, Li, Lin, Sheng, Wu, Zhang, Zheng, Zhuang, Zhuang, Gonzalez, Stoica, and Xing}]{vicuna2023}
Wei-Lin Chiang, Zhuohan Li, Zi~Lin, Ying Sheng, Zhanghao Wu, Hao Zhang, Lianmin Zheng, Siyuan Zhuang, Yonghao Zhuang, Joseph~E. Gonzalez, Ion Stoica, and Eric~P. Xing. 2023.
\newblock \href {https://lmsys.org/blog/2023-03-30-vicuna/} {Vicuna: An open-source chatbot impressing gpt-4 with 90\%* chatgpt quality}.

\bibitem[{Covington and McFall(2010)}]{covington2010cutting}
Michael~A Covington and Joe~D McFall. 2010.
\newblock Cutting the gordian knot: The moving-average type--token ratio (mattr).
\newblock \emph{Journal of quantitative linguistics}, 17(2):94--100.

\bibitem[{Dai et~al.(2023)Dai, Li, Li, Tiong, Zhao, Wang, Li, Fung, and Hoi}]{dai2023instructblip}
Wenliang Dai, Junnan Li, Dongxu Li, Anthony Tiong, Junqi Zhao, Weisheng Wang, Boyang Li, Pascale Fung, and Steven Hoi. 2023.
\newblock \href {https://openreview.net/forum?id=vvoWPYqZJA} {Instruct{BLIP}: Towards general-purpose vision-language models with instruction tuning}.
\newblock In \emph{Thirty-seventh Conference on Neural Information Processing Systems}.

\bibitem[{Fu et~al.(2023)Fu, Chen, Shen, Qin, Zhang, Lin, Yang, Zheng, Li, Sun et~al.}]{fu2023mme}
Chaoyou Fu, Peixian Chen, Yunhang Shen, Yulei Qin, Mengdan Zhang, Xu~Lin, Jinrui Yang, Xiawu Zheng, Ke~Li, Xing Sun, et~al. 2023.
\newblock \href {https://arxiv.org/abs/2306.13394} {Mme: A comprehensive evaluation benchmark for multimodal large language models}.
\newblock \emph{ArXiv preprint}, abs/2306.13394.

\bibitem[{Goyal et~al.(2017)Goyal, Khot, Summers{-}Stay, Batra, and Parikh}]{goyal2017making}
Yash Goyal, Tejas Khot, Douglas Summers{-}Stay, Dhruv Batra, and Devi Parikh. 2017.
\newblock \href {https://doi.org/10.1109/CVPR.2017.670} {Making the {V} in {VQA} matter: Elevating the role of image understanding in visual question answering}.
\newblock In \emph{2017 {IEEE} Conference on Computer Vision and Pattern Recognition, {CVPR} 2017, Honolulu, HI, USA, July 21-26, 2017}, pages 6325--6334. {IEEE} Computer Society.

\bibitem[{Guan et~al.(2023)Guan, Liu, Li, Wang, Yacoob, and Zhou}]{guan2023hallusionbench}
Tianrui Guan, Fuxiao Liu, Xiyang Wu Ruiqi Xian~Zongxia Li, Xiaoyu Liu~Xijun Wang, Lichang Chen Furong Huang~Yaser Yacoob, and Dinesh Manocha~Tianyi Zhou. 2023.
\newblock Hallusionbench: An advanced diagnostic suite for entangled language hallucination \& visual illusion in large vision-language models.
\newblock \emph{arXiv e-prints}, pages arXiv--2310.

\bibitem[{Gupta et~al.(2022)Gupta, Li, Kortylewski, Zhang, Li, and Yuille}]{gupta2022swapmix}
Vipul Gupta, Zhuowan Li, Adam Kortylewski, Chenyu Zhang, Yingwei Li, and Alan~L. Yuille. 2022.
\newblock \href {https://doi.org/10.1109/CVPR52688.2022.00502} {Swapmix: Diagnosing and regularizing the over-reliance on visual context in visual question answering}.
\newblock In \emph{{IEEE/CVF} Conference on Computer Vision and Pattern Recognition, {CVPR} 2022, New Orleans, LA, USA, June 18-24, 2022}, pages 5068--5078. {IEEE}.

\bibitem[{Hu et~al.(2023)Hu, Yao, Wang, Wang, Pan, Chen, Yu, Wu, Zhao, Zhang et~al.}]{viscpm}
Jinyi Hu, Yuan Yao, Chongyi Wang, Shan Wang, Yinxu Pan, Qianyu Chen, Tianyu Yu, Hanghao Wu, Yue Zhao, Haoye Zhang, et~al. 2023.
\newblock Large multilingual models pivot zero-shot multimodal learning across languages.
\newblock \emph{arXiv preprint arXiv:2308.12038}.

\bibitem[{Jain et~al.(2021)Jain, Kothyari, Kumar, Jyothi, Ramakrishnan, and Chakrabarti}]{jain2021select}
Aman Jain, Mayank Kothyari, Vishwajeet Kumar, Preethi Jyothi, Ganesh Ramakrishnan, and Soumen Chakrabarti. 2021.
\newblock Select, substitute, search: A new benchmark for knowledge-augmented visual question answering.
\newblock In \emph{Proceedings of the 44th International ACM SIGIR Conference on Research and Development in Information Retrieval}, pages 2491--2498.

\bibitem[{Karpukhin et~al.(2020)Karpukhin, Oguz, Min, Lewis, Wu, Edunov, Chen, and Yih}]{karpukhin2020dense}
Vladimir Karpukhin, Barlas Oguz, Sewon Min, Patrick Lewis, Ledell Wu, Sergey Edunov, Danqi Chen, and Wen-tau Yih. 2020.
\newblock \href {https://doi.org/10.18653/v1/2020.emnlp-main.550} {Dense passage retrieval for open-domain question answering}.
\newblock In \emph{Proceedings of the 2020 Conference on Empirical Methods in Natural Language Processing (EMNLP)}, pages 6769--6781, Online. Association for Computational Linguistics.

\bibitem[{Khandelwal et~al.(2020)Khandelwal, Levy, Jurafsky, Zettlemoyer, and Lewis}]{khandelwal2019generalization}
Urvashi Khandelwal, Omer Levy, Dan Jurafsky, Luke Zettlemoyer, and Mike Lewis. 2020.
\newblock \href {https://openreview.net/forum?id=HklBjCEKvH} {Generalization through memorization: Nearest neighbor language models}.
\newblock In \emph{8th International Conference on Learning Representations, {ICLR} 2020, Addis Ababa, Ethiopia, April 26-30, 2020}. OpenReview.net.

\bibitem[{Liu et~al.(2023{\natexlab{a}})Liu, Li, Wu, and Lee}]{liu2023visual}
Haotian Liu, Chunyuan Li, Qingyang Wu, and Yong~Jae Lee. 2023{\natexlab{a}}.
\newblock \href {https://arxiv.org/abs/2304.08485} {Visual instruction tuning}.
\newblock \emph{ArXiv preprint}, abs/2304.08485.

\bibitem[{Liu et~al.(2023{\natexlab{b}})Liu, Duan, Zhang, Li, Zhang, Zhao, Yuan, Wang, He, Liu et~al.}]{liu2023mmbench}
Yuan Liu, Haodong Duan, Yuanhan Zhang, Bo~Li, Songyang Zhang, Wangbo Zhao, Yike Yuan, Jiaqi Wang, Conghui He, Ziwei Liu, et~al. 2023{\natexlab{b}}.
\newblock \href {https://arxiv.org/abs/2307.06281} {Mmbench: Is your multi-modal model an all-around player?}
\newblock \emph{ArXiv preprint}, abs/2307.06281.

\bibitem[{Lu et~al.(2024)Lu, Qian, Zheng, Fan, Gao, Zhang, Shao, Deng, Fu, Huang et~al.}]{lu2024gpt}
Chaochao Lu, Chen Qian, Guodong Zheng, Hongxing Fan, Hongzhi Gao, Jie Zhang, Jing Shao, Jingyi Deng, Jinlan Fu, Kexin Huang, et~al. 2024.
\newblock From gpt-4 to gemini and beyond: Assessing the landscape of mllms on generalizability, trustworthiness and causality through four modalities.
\newblock \emph{arXiv preprint arXiv:2401.15071}.

\bibitem[{Marino et~al.(2019)Marino, Rastegari, Farhadi, and Mottaghi}]{marino2019ok}
Kenneth Marino, Mohammad Rastegari, Ali Farhadi, and Roozbeh Mottaghi. 2019.
\newblock \href {https://doi.org/10.1109/CVPR.2019.00331} {{OK-VQA:} {A} visual question answering benchmark requiring external knowledge}.
\newblock In \emph{{IEEE} Conference on Computer Vision and Pattern Recognition, {CVPR} 2019, Long Beach, CA, USA, June 16-20, 2019}, pages 3195--3204. Computer Vision Foundation / {IEEE}.

\bibitem[{McCarthy(2005)}]{mccarthy2005assessment}
Philip~M McCarthy. 2005.
\newblock \emph{An assessment of the range and usefulness of lexical diversity measures and the potential of the measure of textual, lexical diversity (MTLD)}.
\newblock Ph.D. thesis, The University of Memphis.

\bibitem[{McCarthy and Jarvis(2010)}]{mccarthy2010mtld}
Philip~M McCarthy and Scott Jarvis. 2010.
\newblock Mtld, vocd-d, and hd-d: A validation study of sophisticated approaches to lexical diversity assessment.
\newblock \emph{Behavior research methods}, 42(2):381--392.

\bibitem[{Niu et~al.(2021)Niu, Tang, Zhang, Lu, Hua, and Wen}]{niu2021counterfactual}
Yulei Niu, Kaihua Tang, Hanwang Zhang, Zhiwu Lu, Xian{-}Sheng Hua, and Ji{-}Rong Wen. 2021.
\newblock \href {https://doi.org/10.1109/CVPR46437.2021.01251} {Counterfactual {VQA:} {A} cause-effect look at language bias}.
\newblock In \emph{{IEEE} Conference on Computer Vision and Pattern Recognition, {CVPR} 2021, virtual, June 19-25, 2021}, pages 12700--12710. Computer Vision Foundation / {IEEE}.

\bibitem[{OpenAI(2023)}]{gpt2023openai}
OpenAI. 2023.
\newblock Gpt-4 technical report.

\bibitem[{Ouyang et~al.(2022)Ouyang, Wu, Jiang, Almeida, Wainwright, Mishkin, Zhang, Agarwal, Slama, Ray et~al.}]{ouyang2022training}
Long Ouyang, Jeffrey Wu, Xu~Jiang, Diogo Almeida, Carroll Wainwright, Pamela Mishkin, Chong Zhang, Sandhini Agarwal, Katarina Slama, Alex Ray, et~al. 2022.
\newblock Training language models to follow instructions with human feedback.
\newblock \emph{Advances in Neural Information Processing Systems}, 35:27730--27744.

\bibitem[{Parcalabescu et~al.(2022)Parcalabescu, Cafagna, Muradjan, Frank, Calixto, and Gatt}]{parcalabescu-etal-2022-valse}
Letitia Parcalabescu, Michele Cafagna, Lilitta Muradjan, Anette Frank, Iacer Calixto, and Albert Gatt. 2022.
\newblock \href {https://doi.org/10.18653/v1/2022.acl-long.567} {{VALSE}: A task-independent benchmark for vision and language models centered on linguistic phenomena}.
\newblock In \emph{Proceedings of the 60th Annual Meeting of the Association for Computational Linguistics (Volume 1: Long Papers)}, pages 8253--8280, Dublin, Ireland. Association for Computational Linguistics.

\bibitem[{Parcalabescu and Frank(2023)}]{parcalabescu-frank-2023-mm}
Letitia Parcalabescu and Anette Frank. 2023.
\newblock \href {https://doi.org/10.18653/v1/2023.acl-long.223} {{MM}-{SHAP}: A performance-agnostic metric for measuring multimodal contributions in vision and language models {\&} tasks}.
\newblock In \emph{Proceedings of the 61st Annual Meeting of the Association for Computational Linguistics (Volume 1: Long Papers)}, pages 4032--4059, Toronto, Canada. Association for Computational Linguistics.

\bibitem[{Pearl(1995)}]{pearl1995causal}
Judea Pearl. 1995.
\newblock Causal diagrams for empirical research.
\newblock \emph{Biometrika}, 82(4):669--688.

\bibitem[{Pearl(2022)}]{pearl2022direct}
Judea Pearl. 2022.
\newblock Direct and indirect effects.
\newblock In \emph{Probabilistic and causal inference: the works of Judea Pearl}, pages 373--392.

\bibitem[{Radford et~al.(2019)Radford, Wu, Child, Luan, Amodei, Sutskever et~al.}]{radford2019language}
Alec Radford, Jeffrey Wu, Rewon Child, David Luan, Dario Amodei, Ilya Sutskever, et~al. 2019.
\newblock Language models are unsupervised multitask learners.
\newblock \emph{OpenAI blog}, 1(8):9.

\bibitem[{Rohrbach et~al.(2018)Rohrbach, Hendricks, Burns, Darrell, and Saenko}]{rohrbach-etal-2018-object}
Anna Rohrbach, Lisa~Anne Hendricks, Kaylee Burns, Trevor Darrell, and Kate Saenko. 2018.
\newblock \href {https://doi.org/10.18653/v1/D18-1437} {Object hallucination in image captioning}.
\newblock In \emph{Proceedings of the 2018 Conference on Empirical Methods in Natural Language Processing}, pages 4035--4045, Brussels, Belgium. Association for Computational Linguistics.

\bibitem[{Schwenk et~al.(2022)Schwenk, Khandelwal, Clark, Marino, and Mottaghi}]{schwenk2022okvqa}
Dustin Schwenk, Apoorv Khandelwal, Christopher Clark, Kenneth Marino, and Roozbeh Mottaghi. 2022.
\newblock A-okvqa: A benchmark for visual question answering using world knowledge.
\newblock In \emph{European Conference on Computer Vision}, pages 146--162. Springer.

\bibitem[{Shen(2022)}]{shen2022lexicalrichness}
Lucas Shen. 2022.
\newblock Lexicalrichness: A small module to compute textual lexical richness.

\bibitem[{Si et~al.(2022)Si, Meng, Zheng, Lin, Liu, Fu, Cao, Wang, and Zhou}]{si2022language}
Qingyi Si, Fandong Meng, Mingyu Zheng, Zheng Lin, Yuanxin Liu, Peng Fu, Yanan Cao, Weiping Wang, and Jie Zhou. 2022.
\newblock \href {https://aclanthology.org/2022.findings-emnlp.271} {Language prior is not the only shortcut: A benchmark for shortcut learning in {VQA}}.
\newblock In \emph{Findings of the Association for Computational Linguistics: EMNLP 2022}, pages 3698--3712, Abu Dhabi, United Arab Emirates. Association for Computational Linguistics.

\bibitem[{Stolfo et~al.(2023)Stolfo, Jin, Shridhar, Schoelkopf, and Sachan}]{stolfo-etal-2023-causal}
Alessandro Stolfo, Zhijing Jin, Kumar Shridhar, Bernhard Schoelkopf, and Mrinmaya Sachan. 2023.
\newblock \href {https://doi.org/10.18653/v1/2023.acl-long.32} {A causal framework to quantify the robustness of mathematical reasoning with language models}.
\newblock In \emph{Proceedings of the 61st Annual Meeting of the Association for Computational Linguistics (Volume 1: Long Papers)}, pages 545--561, Toronto, Canada. Association for Computational Linguistics.

\bibitem[{Team et~al.(2023)Team, Anil, Borgeaud, Wu, Alayrac, Yu, Soricut, Schalkwyk, Dai, Hauth et~al.}]{team2023gemini}
Gemini Team, Rohan Anil, Sebastian Borgeaud, Yonghui Wu, Jean-Baptiste Alayrac, Jiahui Yu, Radu Soricut, Johan Schalkwyk, Andrew~M Dai, Anja Hauth, et~al. 2023.
\newblock \href {https://arxiv.org/abs/2312.11805} {Gemini: a family of highly capable multimodal models}.
\newblock \emph{ArXiv preprint}, abs/2312.11805.

\bibitem[{Touvron et~al.(2023{\natexlab{a}})Touvron, Lavril, Izacard, Martinet, Lachaux, Lacroix, Rozi{\`{e}}re, Goyal, Hambro, Azhar, Rodriguez, Joulin, Grave, and Lample}]{Hugo2023LLaMA}
Hugo Touvron, Thibaut Lavril, Gautier Izacard, Xavier Martinet, Marie{-}Anne Lachaux, Timoth{\'{e}}e Lacroix, Baptiste Rozi{\`{e}}re, Naman Goyal, Eric Hambro, Faisal Azhar, Aur{\'{e}}lien Rodriguez, Armand Joulin, Edouard Grave, and Guillaume Lample. 2023{\natexlab{a}}.
\newblock \href {https://arxiv.org/abs/2302.13971} {Llama: Open and efficient foundation language models}.
\newblock \emph{ArXiv preprint}, abs/2302.13971.

\bibitem[{Touvron et~al.(2023{\natexlab{b}})Touvron, Lavril, Izacard, Martinet, Lachaux, Lacroix, Rozi{\`e}re, Goyal, Hambro, Azhar et~al.}]{touvron2023llama}
Hugo Touvron, Thibaut Lavril, Gautier Izacard, Xavier Martinet, Marie-Anne Lachaux, Timoth{\'e}e Lacroix, Baptiste Rozi{\`e}re, Naman Goyal, Eric Hambro, Faisal Azhar, et~al. 2023{\natexlab{b}}.
\newblock \href {https://arxiv.org/abs/2302.13971} {Llama: Open and efficient foundation language models}.
\newblock \emph{ArXiv preprint}, abs/2302.13971.

\bibitem[{Wang et~al.(2017)Wang, Wu, Shen, Dick, and Van Den~Hengel}]{wang2017fvqa}
Peng Wang, Qi~Wu, Chunhua Shen, Anthony Dick, and Anton Van Den~Hengel. 2017.
\newblock Fvqa: Fact-based visual question answering.
\newblock \emph{IEEE transactions on pattern analysis and machine intelligence}, 40(10):2413--2427.

\bibitem[{Wang et~al.(2019)Wang, Huang, Jiang, Knight, Ji, Bansal, and Luan}]{wang-etal-2019-paperrobot}
Qingyun Wang, Lifu Huang, Zhiying Jiang, Kevin Knight, Heng Ji, Mohit Bansal, and Yi~Luan. 2019.
\newblock \href {https://doi.org/10.18653/v1/P19-1191} {{P}aper{R}obot: Incremental draft generation of scientific ideas}.
\newblock In \emph{Proceedings of the 57th Annual Meeting of the Association for Computational Linguistics}, pages 1980--1991, Florence, Italy. Association for Computational Linguistics.

\bibitem[{Wang et~al.(2021)Wang, Gao, Zhu, Zhang, Liu, Li, and Tang}]{wang2021kepler}
Xiaozhi Wang, Tianyu Gao, Zhaocheng Zhu, Zhengyan Zhang, Zhiyuan Liu, Juanzi Li, and Jian Tang. 2021.
\newblock \href {https://doi.org/10.1162/tacl_a_00360} {{KEPLER}: A unified model for knowledge embedding and pre-trained language representation}.
\newblock \emph{Transactions of the Association for Computational Linguistics}, 9:176--194.

\bibitem[{Wei et~al.(2022)Wei, Wang, Schuurmans, Bosma, Xia, Chi, Le, Zhou et~al.}]{wei2022chain}
Jason Wei, Xuezhi Wang, Dale Schuurmans, Maarten Bosma, Fei Xia, Ed~Chi, Quoc~V Le, Denny Zhou, et~al. 2022.
\newblock Chain-of-thought prompting elicits reasoning in large language models.
\newblock \emph{Advances in Neural Information Processing Systems}, 35:24824--24837.

\bibitem[{Yang et~al.(2023)Yang, Li, Lin, Wang, Lin, Liu, and Wang}]{yang2023dawn}
Zhengyuan Yang, Linjie Li, Kevin Lin, Jianfeng Wang, Chung-Ching Lin, Zicheng Liu, and Lijuan Wang. 2023.
\newblock \href {https://arxiv.org/abs/2309.17421} {The dawn of lmms: Preliminary explorations with gpt-4v (ision)}.
\newblock \emph{ArXiv preprint}, abs/2309.17421.

\bibitem[{Ye et~al.(2023)Ye, Xu, Xu, Ye, Yan, Zhou, Wang, Hu, Shi, Shi et~al.}]{ye2023mplug}
Qinghao Ye, Haiyang Xu, Guohai Xu, Jiabo Ye, Ming Yan, Yiyang Zhou, Junyang Wang, Anwen Hu, Pengcheng Shi, Yaya Shi, et~al. 2023.
\newblock \href {https://arxiv.org/abs/2304.14178} {mplug-owl: Modularization empowers large language models with multimodality}.
\newblock \emph{ArXiv preprint}, abs/2304.14178.

\bibitem[{Yu et~al.(2024)Yu, Yao, Zhang, He, Han, Cui, Hu, Liu, Zheng, Sun et~al.}]{rlhfv}
Tianyu Yu, Yuan Yao, Haoye Zhang, Taiwen He, Yifeng Han, Ganqu Cui, Jinyi Hu, Zhiyuan Liu, Hai-Tao Zheng, Maosong Sun, et~al. 2024.
\newblock Rlhf-v: Towards trustworthy mllms via behavior alignment from fine-grained correctional human feedback.
\newblock In \emph{Proceedings of the IEEE/CVF Conference on Computer Vision and Pattern Recognition}, pages 13807--13816.

\bibitem[{Zellers et~al.(2019)Zellers, Bisk, Farhadi, and Choi}]{zellers2019recognition}
Rowan Zellers, Yonatan Bisk, Ali Farhadi, and Yejin Choi. 2019.
\newblock \href {https://doi.org/10.1109/CVPR.2019.00688} {From recognition to cognition: Visual commonsense reasoning}.
\newblock In \emph{{IEEE} Conference on Computer Vision and Pattern Recognition, {CVPR} 2019, Long Beach, CA, USA, June 16-20, 2019}, pages 6720--6731. Computer Vision Foundation / {IEEE}.

\bibitem[{Zhu et~al.(2020)Zhu, Mao, Liu, Zhang, Wang, and Zhang}]{zhu2020overcoming}
Xi~Zhu, Zhendong Mao, Chunxiao Liu, Peng Zhang, Bin Wang, and Yongdong Zhang. 2020.
\newblock \href {https://doi.org/10.24963/ijcai.2020/151} {Overcoming language priors with self-supervised learning for visual question answering}.
\newblock In \emph{Proceedings of the Twenty-Ninth International Joint Conference on Artificial Intelligence, {IJCAI} 2020}, pages 1083--1089. ijcai.org.

\bibitem[{Zhu et~al.(2016)Zhu, Groth, Bernstein, and Fei{-}Fei}]{zhu2016visual7w}
Yuke Zhu, Oliver Groth, Michael~S. Bernstein, and Li~Fei{-}Fei. 2016.
\newblock \href {https://doi.org/10.1109/CVPR.2016.540} {Visual7w: Grounded question answering in images}.
\newblock In \emph{2016 {IEEE} Conference on Computer Vision and Pattern Recognition, {CVPR} 2016, Las Vegas, NV, USA, June 27-30, 2016}, pages 4995--5004. {IEEE} Computer Society.

\end{thebibliography}
\clearpage
\appendix
\begin{table}
    \renewcommand
    \arraystretch{1.0}
    \centering
    \setlength{\tabcolsep}{6pt}
    \begin{tabular}{l|c|c|c}
    \toprule
         {\textbf{Dataset}}  & \textbf{\#{I, Q, A}} &\textbf{Len of Q / A}  & \textbf{\# Ent}\\ \midrule
         \texttt{MORE}-train &10,000 &14.3 / 2.1 &1,261 \\
        \ \ \ \ - 2-hop &4,134 &11.6 / 2.0 &886 \\
         \ \ \ \ - 3-hop &5,866 &16.1 / 2.2 &686 \\
         \midrule
         \texttt{MORE}-dev &1,000 &13.8 / 2.3 &118 \\
         \ \ \ \ - 2-hop &548 &12.2 / 2.2 &71 \\
         \ \ \ \ - 3-hop &452 &15.8 / 2.5 &73 \\
         \midrule
         \texttt{MORE}-test &1,000 &13.9 / 2.4 &251 \\
         \ \ \ \ - 2-hop &500 &12.3 / 2.2 &153 \\
         \ \ \ \ - 3-hop &500 &15.6 / 2.6 &143 \\
    \bottomrule
    \end{tabular}
       \caption{\label{tab:dataset_stat} Dataset statistics of different hops.}
\end{table}
\section{Prompt Templates}
\subsection{Question Generation}
\label{app:prompt_question}
\begin{figure*}
\centering  
\includegraphics[width=1.0\textwidth]{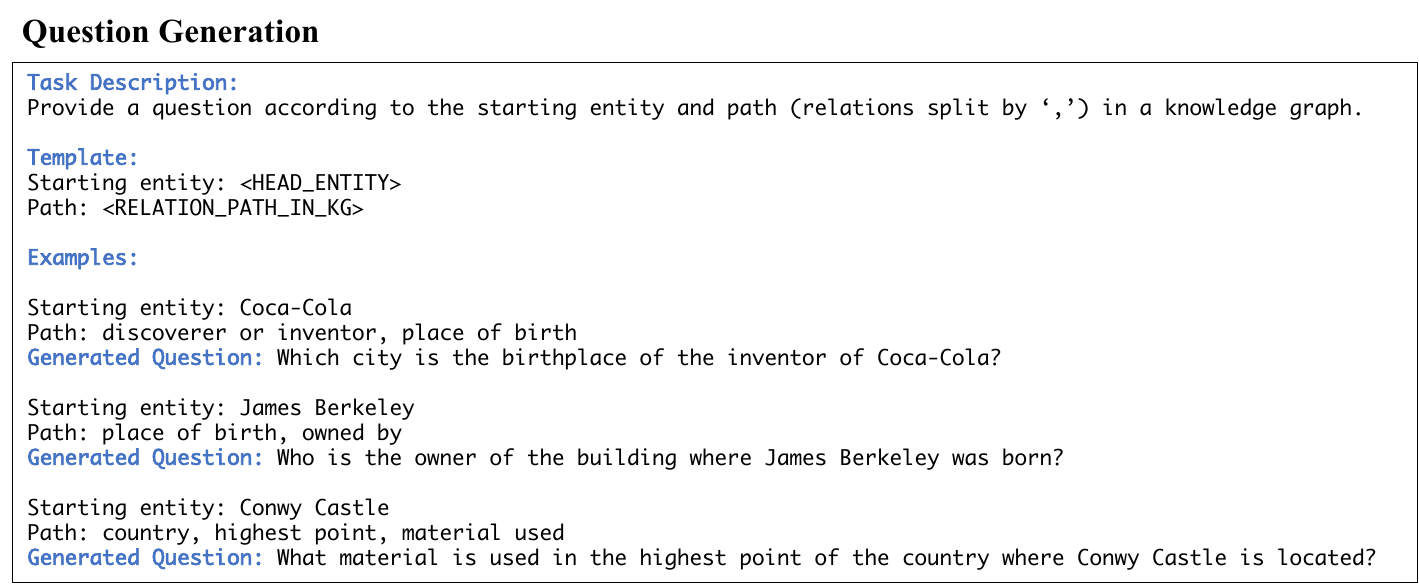}
\caption{Prompt template of multi-hop question generation.}
\label{fig:prompt_question}
\end{figure*}
We present the prompt template for generating language bias options of Section~\ref{subsec:mc_question} in Figure~\ref{fig:prompt_question}.

\subsection{Language Bias Option Generation}
\label{app:prompt_language}
\begin{figure*}
\centering  
\includegraphics[width=1.0\textwidth]{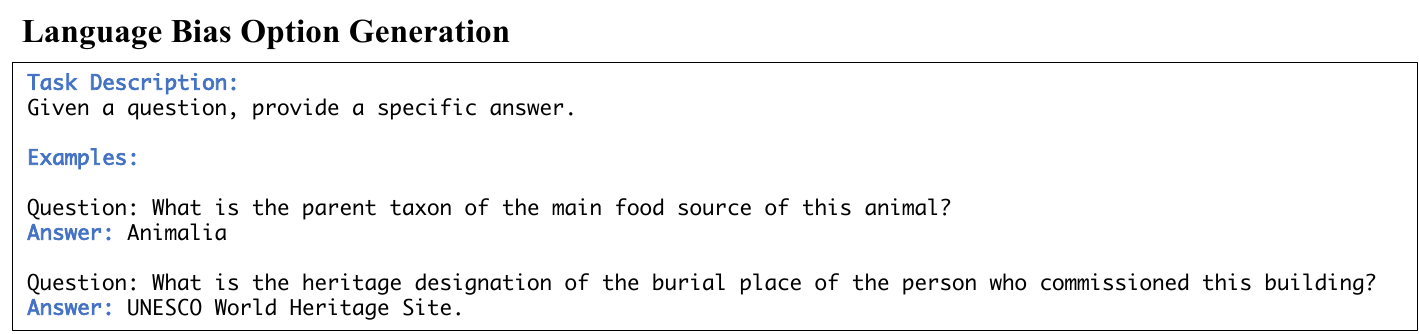}
\caption{Prompt template of language bias option generation.}
\label{fig:prompt_lan}
\end{figure*}
We present the prompt template for generating language bias options of Section~\ref{subsec:mc_question} in Figure~\ref{fig:prompt_lan}.

\section{Question Distribution}

\begin{figure*}
\centering  
\includegraphics[width=1.0\textwidth]{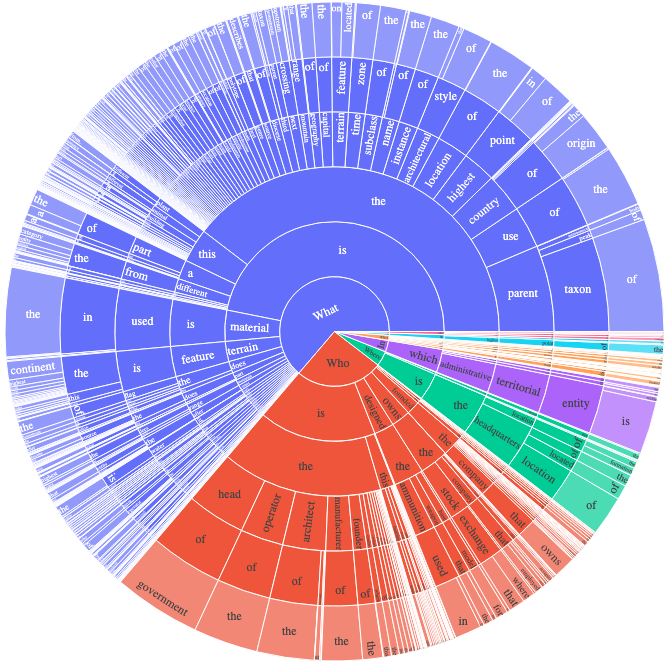}
\caption{Question prefix distribution in \texttt{MORE}. The arc length is proportional to the number of questions containing the word.}
\label{fig:question_distribution}
\end{figure*}

In Figure~\ref{fig:question_distribution}, we categorize the generated questions into distinct types, based on their starting n-grams. The dataset MORE showcases an extensive lexical diversity in the questions generated. This diversity is evidenced by variations in the introductory interrogative words (e.g., ``what'', ``who'', ``where'', etc.), exemplified by phrases like ``What is the...'', ``In which country...'', and more. Such lexical richness is crucial for mitigating the vulnerability of MLLMs to linguistic variations.

\section{Quality Analysis Details}
\subsection{Question Quality}
\label{app:question_quality}
To ensure the quality of the comprising datasets, we analyze the lexical diversity and the fluency of the generated questions, which are useful for conducting a robust evaluation using questions that are linguistically diverse and coherent.
\paragraph{Baselines} We select extensive VQA datasets for comparison, including Visual7W~\cite{zhu2016visual7w}, VQA (v2)~\cite{goyal2017making},  FVQA~\cite{wang2017fvqa}, OK-VQA~\cite{marino2019ok}, S3VQA~\cite{jain2021select}, A-OKVQA~\cite{schwenk2022okvqa},  and INFOSEEK~\cite{chen-etal-2023-pre-trained} (contains both automated generation version and human-annotated version).
\paragraph{Evaluation Metrics}
For lexical diversity, we utilize three metrics that are not dependent on length: moving average type-token ratio (MATTR)~\cite{covington2010cutting}, measure of textual lexical diversity (MTLD)~\cite{mccarthy2005assessment}, and hypergeometric distribution diversity (HDD)~\cite{mccarthy2010mtld}. We average these three metrics for a unified assessment and employ the Lexical-Richness package~\cite{shen2022lexicalrichness} (version 0.5.03) for calculation. For fluency, we employ a pre-trained language model GPT2-large~\cite{radford2019language} with 774M parameters to compute the perplexity of the questions, which is often used as a measure by previous work~\cite{wang-etal-2019-paperrobot, cahyawijaya-etal-2021-indonlg}.

\begin{figure}
\centering  
\includegraphics[width=0.48\textwidth]{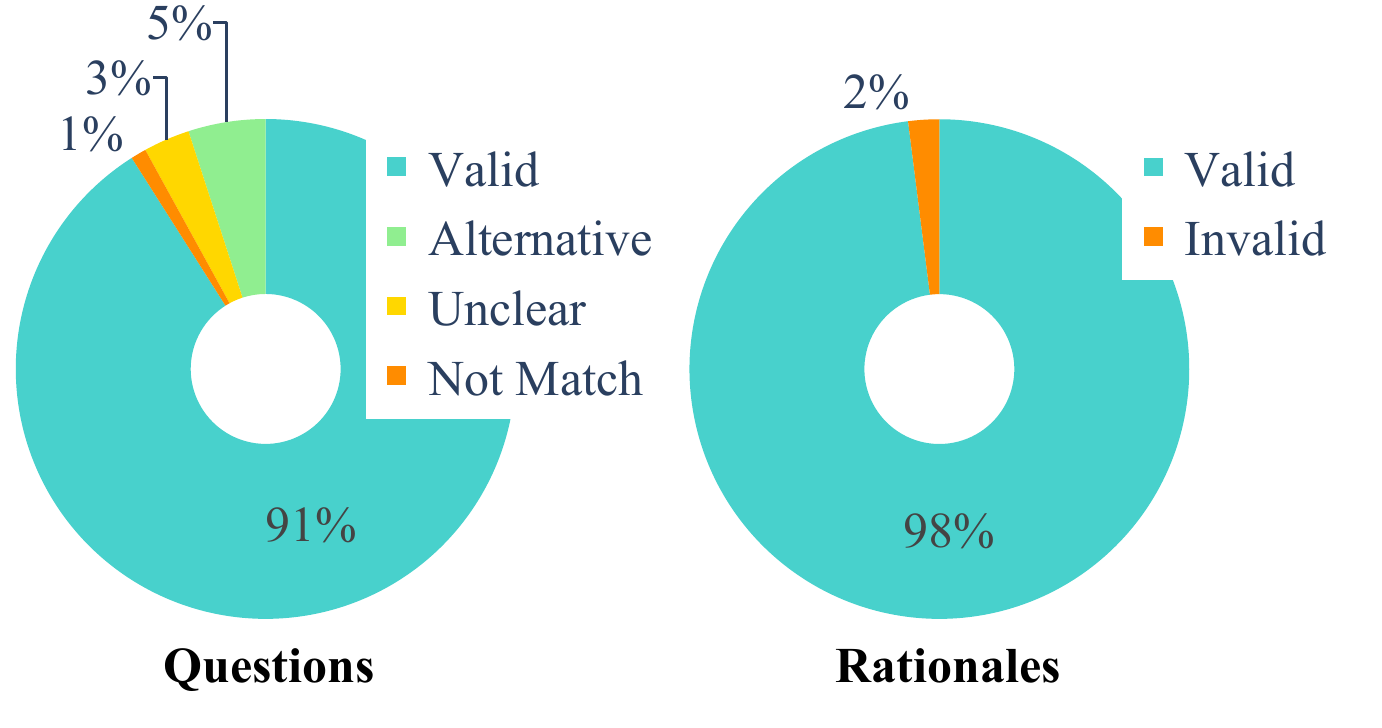}
\caption{Human evaluation results of \texttt{MORE}.}
\label{fig:human_evaluation}
\end{figure}

\subsection{Human Evaluation}
\label{app:human_evaluation}
\paragraph{Questions}
We conduct a human evaluation of 100 questions randomly chosen from the \texttt{MORE} dataset to validate and assess the quality of the generated questions. This evaluation is carried out independently by three human annotators, who are provided with detailed guidelines and illustrative examples before starting the evaluation process. For each question, given the visual context and ground truth answer, we first ask two junior annotators to determine whether: 1) the question is valid, 2) the question allows for an alternative answer, 3) the question does not match the answer, or 4) the question is unclear or ambiguous. If the choices of the two annotators are inconsistent, a senior annotator checks the answers and makes the final decision. The average inter-annotator agreement is 88.6\%~(Cohen's kappa).
\paragraph{Rationales}
We also conduct a human evaluation of the causal rationales, following the same procedure as described above. The difference is that here we provide annotators with only two options: to assess whether the generated rationales are valid.

As shown in Figure~\ref{fig:human_evaluation}, the results are encouraging, with 91\% questions and 98\% rationales being classified as valid by the annotators, further demonstrating the quality of our datasets.

\section{Baselines}
\label{app:baselines}
For open-source MLLMs, we consider the following baselines:

1) InstructBLIP~\cite{dai2023instructblip}, an instruction-tuned version of BLIP-2 on various tasks including VQA. We employ its InstructBLIP-Vicuna~\cite{vicuna2023}-13B variant.

2) mPLUG-Owl~\cite{ye2023mplug}, which proposes a new two-stage training method for aligning images and text. We employ its mPLUG-Owl-Llama~\cite{Hugo2023LLaMA}-7B variant.

3) LLaVA~\cite{liu2023visual}, which translates images into texts of captions and bounding boxes, and prompts GPT-4 to generate a multimodal instruct-tuning dataset in the context of seed examples. We employ its LLaVA-Llama~\cite{Hugo2023LLaMA}-13B variant.

4) Qwen-VL~\cite{bai2023qwen}, which builds upon Qwen~\cite{bai2023qwent} and employ multi-stage training pipeline. Qwen-VL facilitates grounding and text comprehension by aligning image-caption-box tuples. It processes inputs of images, text, and bounding boxes, and produces corresponding text and bounding boxes as outputs.

\section{Details of the \texttt{CAVE} Framework}
\label{app:cave}
In this section, we first present the overall framework as shown in Figure~\ref{fig:cave}, and then we will go over each part of it in detail.

\subsection{Overall Framework}
\begin{figure*}
\centering 
\includegraphics[width=1.0\textwidth]{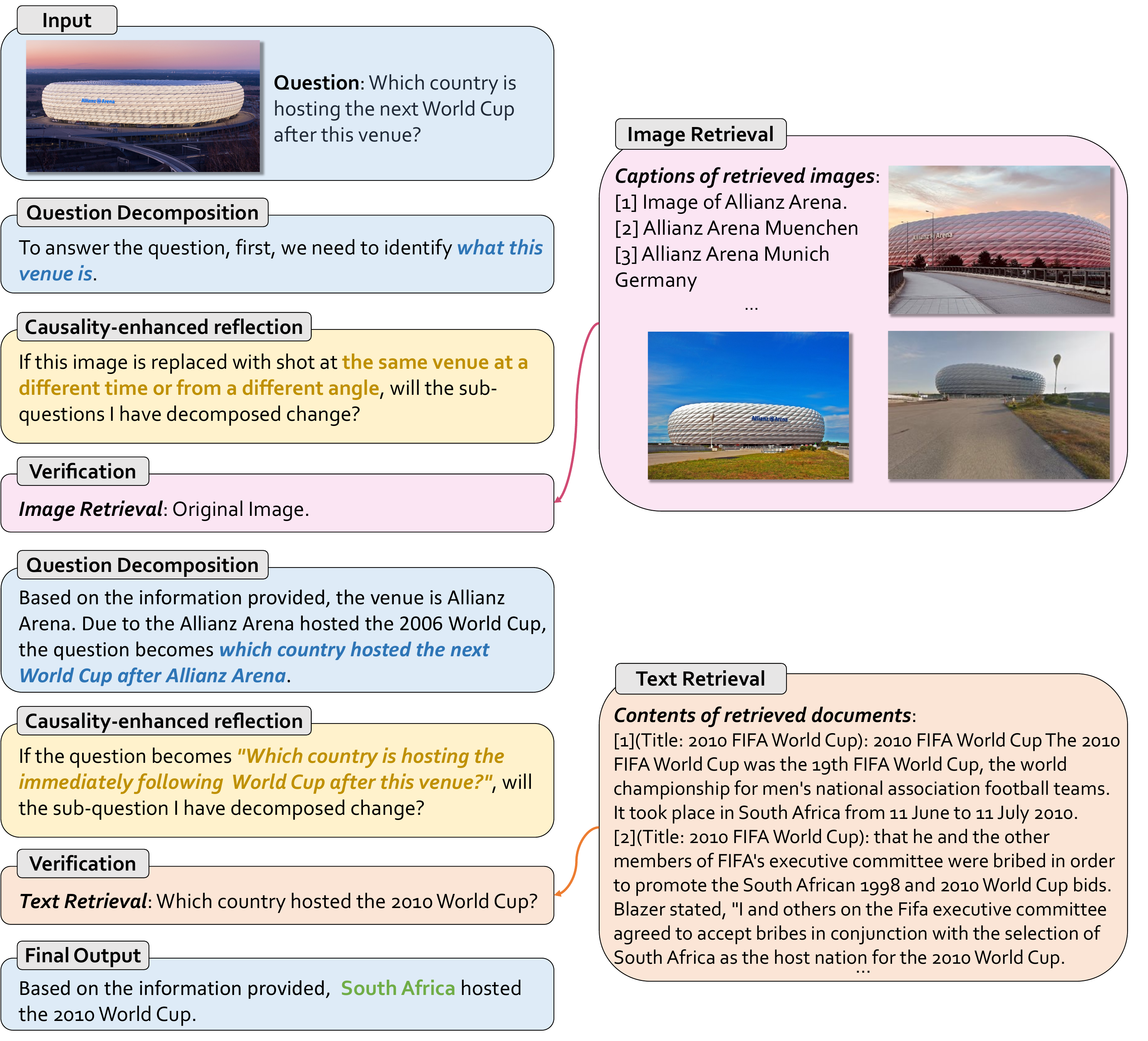} 
\caption{An overview of our proposed \texttt{CAVE} Framework.} 
\label{fig:cave} 
\end{figure*}

Given a question $Q$ and an image $I$, the VQA task demands the system to return an output $A$ that concisely answers the question.
As shown in Figure~\ref{fig:cave}, we first initialize a question decomposer $D$ to analyze $Q$ and break it down into manageable subquestions.
For each subquestion, we introduce a causality-enhanced reasoner $C$ to evaluate the correctness of the decomposed subquestion.
Then, we employ a verifier $V$ to confirm the accuracy of the original answer to each subquestion.
The generation verifier typically involves active information-seeking and answer-verification, which acquires the necessary context or information needed to investigate and revise the answers.
This includes two optional operations: image retrieval to seek images similar to $I$ and determine their titles, or text retrieval with a specific query to fetch pertinent documents and summarize their content.
This iterative process of answering and verifying will continue until we resolve each subquestion.
Finally, the verified answer $A$ is output following the aforementioned reasoning process and retrieved information.

\subsection{Question Decomposer}
For a given VQA problem, MLLMs may simply exploit a spurious shortcut to make predictions. In order to alleviate this issue, motivated by Chain-of-Thought reasoning~\cite{wei2022chain}, we encourage MLLMs to decompose the question $Q$ before outputting the answer, 
so as to gradually solve a complex question that requires multi-hop reasoning. As shown in Figure~\ref{fig:cave}, for the question ``\emph{Which country hosted the next World Cup after this venue?}'', our decomposer breaks it down into two subquestions :
\begin{itemize}
\item [1.] ``\emph{What this venue is}?
\item [2.] ``\emph{Which country hosted the next World Cup after Allianz Arena}?
\end{itemize}
Such decomposition will explicitly constrain the model from comprehending and extracting the truth semantics of the question, thus avoiding simply exploring a spurious path to give the answer.

\subsection{Causality-enhanced Reasoner}
The accuracy of sub-question decomposition plays a crucial role in the subsequent reasoning process. To address this, we introduce causality-enhanced self-reflection, which explicitly guides the model in validating the sub-question it has decomposed.
Specifically, based on the proposed causal framework, we guide the model to assess whether the decomposed subquestion changes under two conditions: 1) rephrasing the question text, or 2) replacing the image with another captured by the same entity from a different angle or at a different time. If the model identifies that the decomposed subquestion needs to be altered, this indicates that its initial understanding is influenced by irrelevant factors or biases, failing to capture the true semantics. In such cases, we will enforce a re-decomposition of the question.

\subsection{Verifier}
Some works have found that vision illusion and language hallucination may appear in the process of MLLMs' response generation~\cite{guan2023hallusionbench, yang2023dawn}. To alleviate this issue, we adopt the retrieval-augmented generation approaches~\cite{khandelwal2019generalization, chen2022murag}. Specifically, we consider two different retrieval ways for the verifier to choose during each verification step: image retrieval and text retrieval.

\paragraph{Image Retrieval}
Although our framework is applicable to any image retrieval method,
in this paper, we mainly utilize Google Image Search to obtain a broad range of information related to the image as provided by Google Lens API\footnote{Web interface available at https://images.google.com.}. This information encompasses various details, such as knowledge graph entities and captions of analogous or identical images. The availability of these details can vary based on the image input provided to Google Image Search. 
Then, the verifier gleans relevant information from captions associated with visually similar images, so as to verify the original answer and conduct the next round of reasoning.

\paragraph{Text Retrieval}
Similarly, our framework is applicable to any text retrieval method, we explore a simple, off-the-shelf dense retriever for Wikipedia,  GTR~\cite{karpukhin2020dense},  as our text retriever.
First, the verifier constructs a query to perform text retrieval according to the currently generated context, and then the query is input into a GTR model to get related document titles and contents. Finally, the verifier will fetch pertinent documents and summarize their content to verify the immediate answer.

\subsection{Final Verified Answer}
Finally, the improved response that takes verification into account is generated. This is executed by a final prompt where the context takes into account all of the previous reasoning steps, the baseline response, and the verification question-answer pairs, so that the corrections can take place. 

\section{Case Study}
\label{app:case}
\begin{figure*}
\centering  
\includegraphics[width=0.8\textwidth]{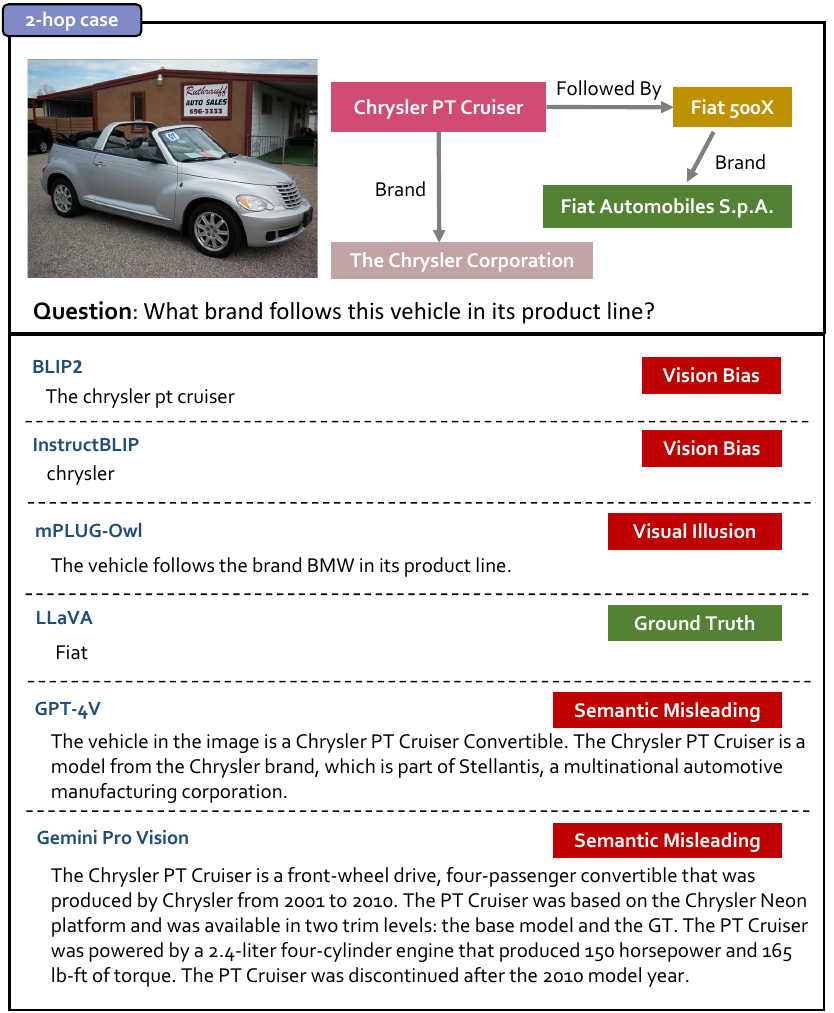}
\caption{Two-hop case in the dev set of \texttt{MORE}. The wrong answers are marked in red and the correct is in green.}
\label{fig:case_two_hop}
\end{figure*}

\begin{figure*}
\centering  
\includegraphics[width=0.8\textwidth]{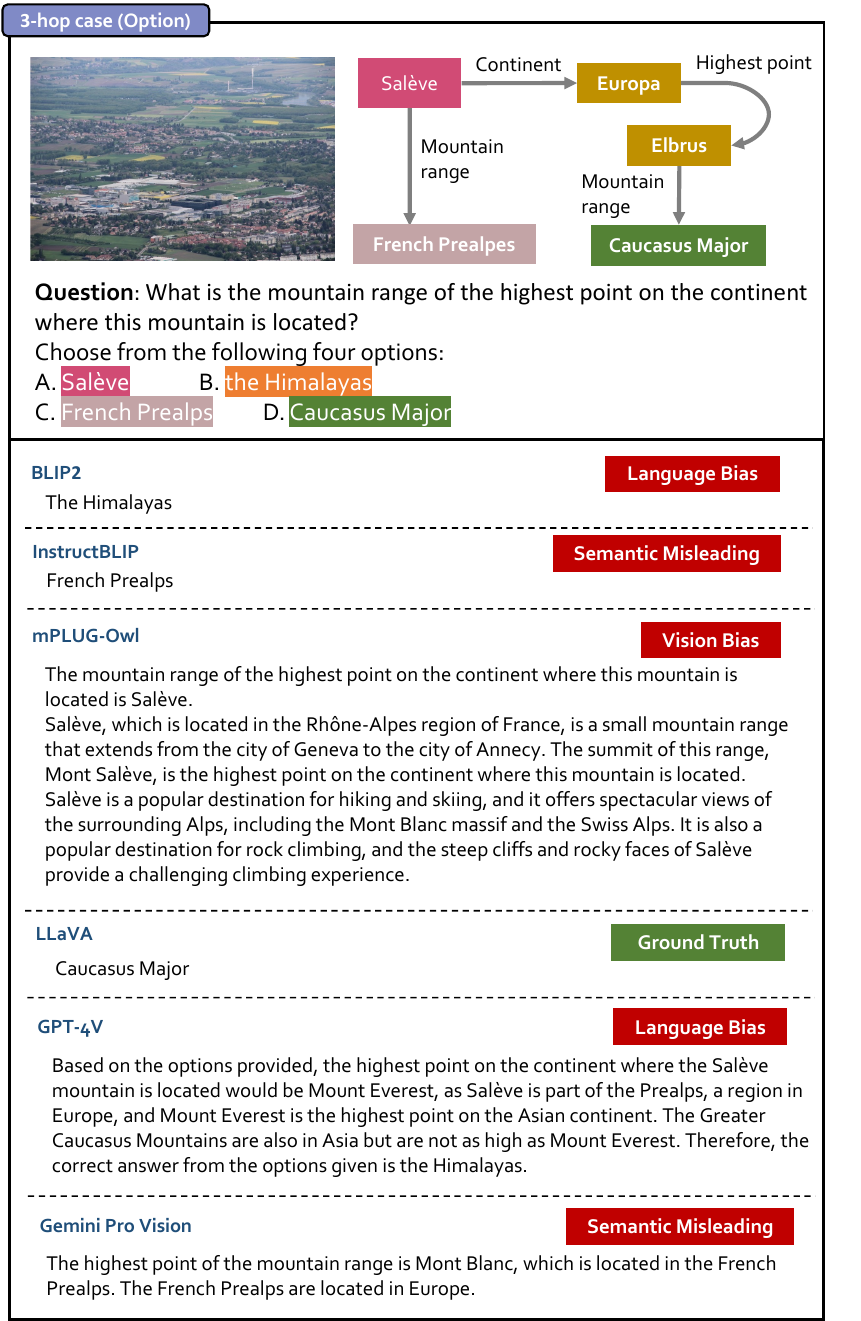}
\caption{Three-hop case (multi-choice setting) in the dev set of \texttt{MORE}. The wrong answers are marked in red and the correct is in green.}
\label{fig:case_three_hop_opt}
\end{figure*}

We conduct a case study on the development set of \texttt{MORE} in Figure~\ref{fig:case_two_hop}$ \sim $\ref{fig:case_three_hop_opt}, including both the ``Open-ended'' and ``Multi-Choice'' settings.

\begin{figure*}
\centering  
\includegraphics[width=0.8\textwidth]{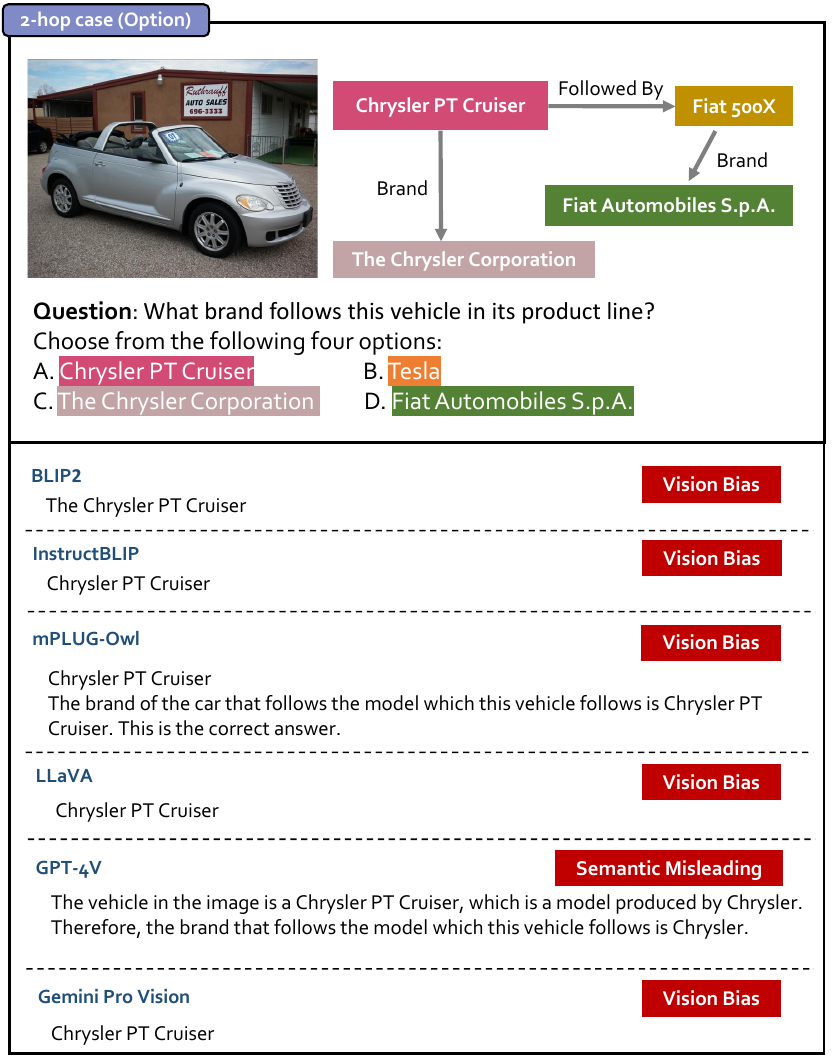}
\caption{Two-hop case (multi-choice setting) in the dev set of \texttt{MORE}. The wrong answers are marked in red.}
\label{fig:case_two_hop_opt}
\end{figure*}

\begin{figure*}
\centering  
\includegraphics[width=0.8\textwidth]{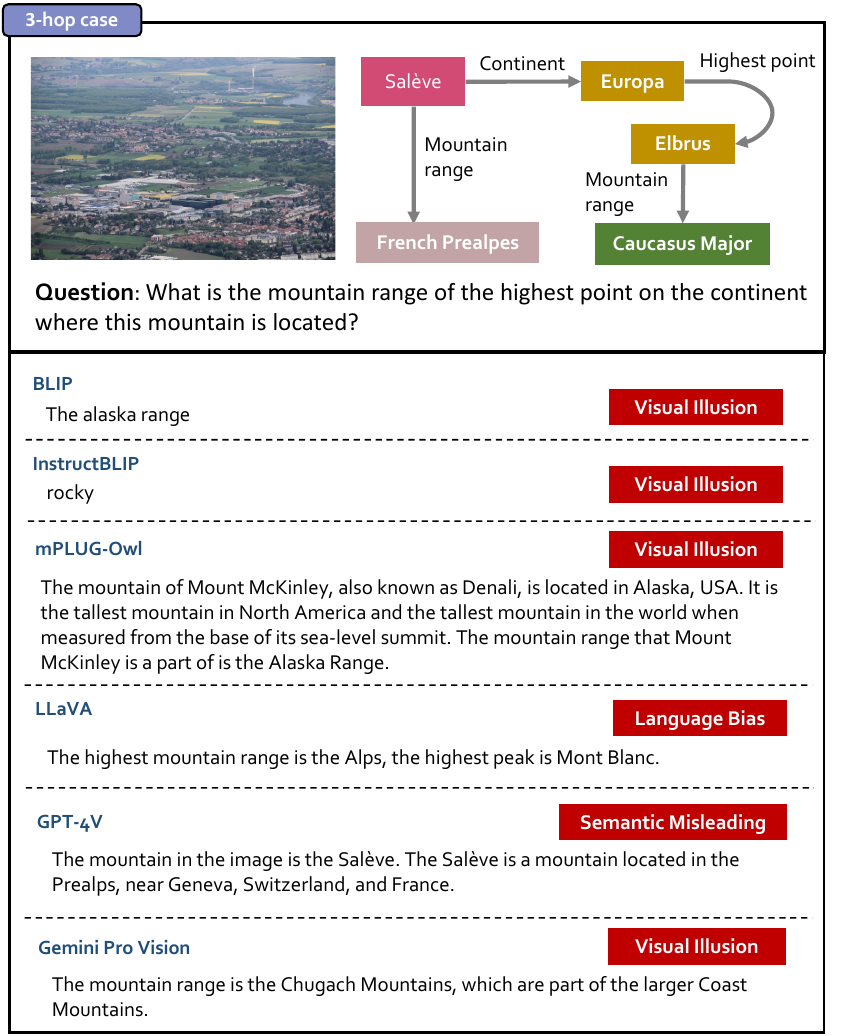}
\caption{Three-hop case in the dev set of \texttt{MORE}. The wrong answers are marked in red.}
\label{fig:case_three_hop}
\end{figure*}

\end{document}